\documentclass[11pt]{article}

\usepackage[preprint]{acl}

\usepackage[most]{tcolorbox}

\usepackage{times}
\usepackage{latexsym}
\usepackage{paralist, tabularx}
\usepackage{booktabs}
\usepackage{amsmath}
\usepackage{tablefootnote}
\usepackage{todonotes}
\usepackage{enumitem}
\usepackage{array, booktabs, makecell, multirow}
\usepackage{listings}
\usepackage{adjustbox}
\usepackage[T1]{fontenc}

\usepackage[utf8]{inputenc}

\usepackage{microtype}

\usepackage{inconsolata}

\usepackage{graphicx}
\usepackage{xcolor}    
\usepackage{pifont}    
\usepackage{amssymb}
\usepackage{bbding}
\newcommand{\cmark}{\textcolor{green}{\checkmark}}
\newcommand{\xmark}{\textcolor{red}{\ding{55}}}
\usepackage{tablefootnote}

\usepackage{hyperref}
\usepackage[T1]{fontenc}
\usepackage[utf8]{inputenc}
\usepackage{microtype}
\usepackage{inconsolata}
\usepackage{graphicx}
\usepackage{xcolor}
\usepackage{amsmath}
\usepackage{booktabs}
\usepackage{multicol}
\usepackage{multirow}
\usepackage{capt-of}
\usepackage{fontawesome5}
\usepackage{newfloat}
\usepackage[most]{tcolorbox}
\usepackage{caption}
\usepackage{subcaption}
\usepackage{enumitem}
\usepackage{tablefootnote}
\usepackage{float}
\usepackage{titlesec}
\usepackage{enumitem}
\usepackage{fancyvrb}
\usepackage{adjustbox}
\usepackage{varwidth}
\usepackage{placeins}
\usepackage[noabbrev,capitalize,nameinlink]{cleveref}

\newcommand{\blue}[1]{\textcolor{black}{#1}}

\title{AICD Bench: A Challenging Benchmark for AI-Generated Code Detection}

\author{
  Daniil Orel$^{1}$,
  Dilshod Azizov$^{1}$,
  Indraneil Paul$^{2}$,
  Yuxia Wang$^{3}$,\\
  \textbf{Iryna Gurevych$^{1,2}$},
  \textbf{Preslav Nakov$^{1}$} \\
  \\
  $^{1}$Mohamed bin Zayed University of Artificial Intelligence (MBZUAI), UAE \\
  $^{2}$Ubiquitous Knowledge Processing Lab (UKP Lab), Department of Computer Science, \\
  TU Darmstadt and National Research Center for Applied Cybersecurity ATHENE, Germany \\
  $^{3}$ INSAIT, Sofia University ``St. Kliment Ohridski'', Bulgaria
  \\
  \texttt{\{name.surname\}@mbzuai.ac.ae; \{name.surname\}@tu-darmstadt.de}
}

\begin{document}
\maketitle
\begin{abstract}
Large language models (LLMs) are increasingly capable of generating functional source code, raising concerns about authorship, accountability, and security. While detecting AI-generated code is critical, existing datasets and benchmarks are narrow, typically limited to binary human-machine classification under in-distribution settings. To bridge this gap, we introduce \emph{AICD Bench}, the most comprehensive benchmark for AI-generated code detection. It spans \emph{2M examples}, \emph{77 models} across \emph{11 families}, and \emph{9 programming languages}, including recent reasoning models. Beyond scale, AICD Bench introduces three realistic detection tasks: (\emph{i})~\emph{Robust Binary Classification} under distribution shifts in language and domain, (\emph{ii})~\emph{Model Family Attribution}, grouping generators by architectural lineage, and (\emph{iii})~\emph{Fine-Grained Human-Machine Classification} across human, machine, hybrid, and adversarial code. Extensive evaluation on neural and classical detectors shows that performance remains far below practical usability, particularly under distribution shift and for hybrid or adversarial code. We release AICD Bench as a \emph{unified, challenging evaluation suite} to drive the next generation of robust approaches for AI-generated code detection. The data and the code are available at \url{https://huggingface.co/AICD-bench}.
\end{abstract}

\section{Introduction}
\label{sec:intro}
With the rapid advances of LLMs, the field of software engineering has also changed. Today, the generation of syntactically and semantically correct functional code can be done at scale. LLMs can not only solve challenging problems from platforms like LeetCode\footnote{\href{https://leetcode.com/}{https://leetcode.com/}}, but also write production-ready code, find and fix bugs, and write unit tests~\cite{deepseekcoder,codellama,qwen3technicalreport,Tufano2020UnitTC}.
Thus, the need to detect such code reliably has emerged as a critical challenge.

Detecting whether a piece of code was written by a human or generated by an LLM is no longer a niche problem and is central to ensuring academic integrity~\cite{salim-etal-2024-impeding, codex_on_exams}, preventing plagiarism~\cite{park2025detectionllmparaphrasedcodeidentification}, and mitigating security risks~\cite{vulnerabilities} associated with undetected synthetic code.

In response to this demand, research on AI-generated code detection has grown rapidly, with numerous methods proposed in recent years. However, this progress has been accompanied by fragmented evaluation: most studies introduce not only a new detection model, but also a dataset tailored to a specific experimental setting. Such datasets typically cover narrow configurations, for example, a small number of programming languages (\emph{e.g.},~only Python, C++, and Java), a limited set of LLMs (\emph{e.g.},~API-based models only), and non-reasoning generation patterns, resulting in highly domain-specific conclusions. While impressive performance is often reported under these controlled conditions, such results provide limited insight into real-world generalization. The core limitation is \emph{the absence of a unified, comprehensive benchmark} capable of evaluating detection methods across multiple dimensions of variation.

Existing evaluations predominantly report in-domain accuracy, neglecting out-of-distribution (OOD) generalization, which is a cornerstone of real-world deployability. Even when OOD evaluation is attempted, it typically considers only one dimension at a time: either changing the programming language (\emph{e.g.}, from Python to Java) or changing the purpose of the code (\emph{e.g.}, from algorithmic problem solving to real-world deployable code) or varying the generator model, but rarely both simultaneously. Such conditions yield optimistic results and do not reflect 
\emph{realistic deployment}, where detectors must handle unseen models, 
programming languages, domains, and code written in an adversarial way. 

To overcome these challenges, we present \emph{AICD Bench}, a \emph{new large-scale benchmark} for AI-generated code detection. 
Unlike prior resources, \emph{AICD Bench} systematically evaluates detectors across multiple axes of variation, including generator families, programming languages, and adversarial strategies. 
It introduces \emph{three tasks} designed to capture increasingly realistic challenges: 
(\emph{i})~\emph{Robust Binary Classification} under language and domain shifts, 
(\emph{ii})~\emph{Model Family Attribution}, grouping generators by architectural lineage, and (\emph{iii})~\emph{Fine-Grained Human-Machine Classification}, distinguishing human, machine, hybrid and adversarial code.

Our contributions are as follows:
\begin{enumerate}
    \item We release \emph{AICD Bench} a comprehensive 2M sample benchmark that spans 77 generators and 9 programming languages, subsuming previous datasets, and provide standardized splits, protocols, and evaluation scripts to enable reproducible research.
    \item We define novel tasks that go beyond binary classification, including model-family attribution and fine-grained classification.
    \item We conduct extensive evaluations of classical and deep learning-based baselines, showing that current methods generalize poorly in out-of-distribution settings.
\end{enumerate}

\section{Related Work}
\label{sec:related-work}
\subsection{Authorship Attribution}
With the widespread use of LLMs, recognizing the author of a piece of work has become an important problem. It can be viewed from multiple perspectives: as a multiway classification problem, where each output has to be paired with an author from a candidate list (which could include LLMs), or as a binary classification problem, e.g.,~``\emph{Is it written by a human or an LLM?}'', ``\emph{Do these outputs share the same author?}'', etc.~\cite{uchendu-etal-2020-authorship}. Previous research in both text and code domains has focused on the first perspective. For example, \citet{wang-etal-2024-m4gt} and \citet{codet} address authorship attribution as a multi-class classification problem, where each class corresponds to a specific LLM. 
In addition to supervised approaches, unsupervised methods have been proposed for authorship identification, including Uniform Information Density~\cite{venkatraman-etal-2024-gpt} and stylistic representation learning~\cite{DBLP:conf/iclr/SotoKKCBA24}. 

However, most existing methods aim to identify the signature of an individual model. As the number of deployed LLMs continues to grow, fine-grained attribution at the level of individual models becomes increasingly impractical. A more scalable alternative is attribution at the level of \emph{model families}, defined as groups of related models that share a common architecture and training philosophy but differ in size, capabilities, or specialization. In \emph{AICD Bench}, we introduce a dedicated task that targets attribution of a given code sample to its underlying model family.

\subsection{Benchmarks for AI-Generated Code Detection}
AI-generated code detection largely follows the trajectory of AI-generated text detection, where progress has been driven by benchmarks, such as RAID~\citep{raid}, MULTITUDE~\citep{multitude}, and MAGE~\citep{mage}, and robust detection models developed through shared tasks~\citep{wang-etal-2025-genai}. Early work on code detection relied on small-scale benchmarks: \citet{gptsniffer} introduced a 7K-sample dataset and trained a CodeBERT-based classifier~\cite{codebert}, marking one of the first systematic approaches. More recently, \citet{gptsensor} proposed a benchmark of 1.1M Java and Python code samples (550K human–AI pairs) and trained UnixCoder~\cite{unixcoder} with a contrastive objective, achieving improved detection performance.

A further step toward increased diversity was taken by \citet{codet}, who released a dataset of nearly 500K code samples across C++, Java, and Python generated by five compact Code-LMs. Beyond expanding language and model coverage, they introduced a hybrid authorship identification task to distinguish human-written, machine-generated, and human--machine co-authored code. This partially addressed the need for evaluations beyond binary distinctions. More recently, the \texttt{Droid} framework~\cite{droid} extended this line of work by proposing a detection pipeline and introducing an adversarial setting in which LLMs are trained to evade detection, increasing realism and difficulty.

Despite these advances, existing resources fall short of constituting a benchmark, as they lack standardized tasks, splits, and evaluation protocols. The prominent exception is CodeMirage~\cite{codemirage}, which presents a dataset of 210K samples spanning 10 programming languages and generated by ten code LLMs.

CodeMirage evaluates ten diverse detectors, including zero-shot, pre-trained, fine-tuned, and embedding-based approaches, making it the only AI-generated code detection benchmark to date. Nevertheless, it focuses exclusively on a single, simplified task: binary classification of human-written versus AI-generated code. As reported in their results, fine-tuned models such as CodeT5+ achieve F1-scores exceeding 80\% on unseen generators and paraphrased inputs, suggesting that the benchmark may already be saturated.

This situation exposes a critical gap: the lack of benchmarks that systematically evaluate detection methods across multiple dimensions of complexity, including cross-model, cross-language, cross-domain, and hybrid or adversarial generation settings, while reflecting realistic deployment scenarios. Addressing this gap is the primary motivation behind \emph{AICD Bench}.

\section{Motivation Behind the Task Design}
\label{sec:rationale}

The task design of \emph{AICD Bench} is guided by two complementary goals: faithfully capturing the practical challenges of AI-generated code detection and addressing key limitations of existing benchmarks. As discussed in \Cref{sec:related-work}, prior datasets and evaluations have largely centered on binary classification: human-written vs. machine-generated code. While this task is fundamental, it fails to capture the increasingly diverse, hybrid, and adversarial ways in which LLMs are used in practice. In order to bridge this gap, we deliberately design a set of three tasks that progressively increase in complexity and realism.

\paragraph{Task 1: Robust Binary Classification.}
Binary detection (human-written vs. machine-generated) remains a core requirement for applications such as academic integrity enforcement, intellectual property protection, and software security. However, most existing evaluations have focused on in-distribution settings, which can substantially overestimate real-world performance. In practice, detectors must generalize to unseen programming languages and application domains, where stylistic conventions and structural patterns differ markedly. To reflect this reality, Task~1 explicitly partitions evaluation into progressively more challenging out-of-distribution splits. This design enables a systematic assessment of robustness under realistic distribution shifts, yielding a more stringent and informative evaluation than prior benchmarks.

\vspace{6pt}
\paragraph{Task 2: Model Family Attribution.}
Determining whether code is AI-generated is only a first step; understanding \emph{which kind} of model produced it is essential for applications such as corporate intelligence and intellectual property protection. Rather than attributing outputs to individual models, which has become increasingly impractical as the number of code-generating LLMs grows, Task~2 focuses on \emph{model family attribution}. By grouping generators according to shared architectures, training regimes, and design principles, this task provides actionable attribution that balances scalability with informative stylistic characterization.

\paragraph{Task 3: Fine-Grained Human--Machine Classification.}
A simple binary distinction between human- and machine-generated code overlooks the increasingly hybrid and adversarial nature of LLM-assisted development. In practice, developers frequently co-author code with LLMs, while alignment and adversarial training regimes (\emph{e.g.},~RLHF) deliberately shape model outputs to resemble human-written code. Consequently, a realistic benchmark must distinguish between \emph{fully human-written}, \emph{fully AI-generated}, \emph{hybrid}, and \emph{adversarial} code. Task~3 addresses this need by framing detection as a multi-way classification problem that mirrors real-world deployment scenarios, in which detectors must operate reliably under ambiguity and intentional obfuscation.

\section{Data}

Our work builds upon \texttt{Droid}~\cite{droid}, which covers 7 programming languages, 43 LLMs, and nearly 1M code samples, making it one of the largest existing resources for AI-generated code detection. While comprehensive, \texttt{DroidCollection} lacks recent reasoning-oriented models and does not cover several widely used programming languages. To address this, we extended the dataset by adding two additional languages (PHP and Rust) and incorporating a diverse set of new LLMs at a scale comparable to the original collection, explicitly including reasoning models. All newly generated samples follow the original \texttt{Droid} generation protocol to ensure consistency. We generated \blue{500K samples} using inverse-instruction prompting based on \emph{StarCoderData}~\cite{li2023starcodersourceyou} and \emph{CodeNet}~\cite{codenet}, and performed docstring- and comment-conditioned generation using data from \emph{The Vault}~\cite{thevault} to increase diversity and coverage.

\begin{table}[tbh]
\centering
\small
\renewcommand{\arraystretch}{1.2}
\setlength{\tabcolsep}{8pt}
\resizebox{0.48\textwidth}{!}{%
\begin{tabular}{@{}c c c c@{}}
\toprule
\textbf{Task} & \textbf{Split} & \textbf{\# Samples} & \textbf{Languages}  \\
\midrule
\multirow{3}{*}{Task 1} 
  & Train      & 500K  & \makecell[l]{Python, Java, C++} \\
  & Validation & 100K  & \makecell[l]{Python, Java, C++} \\
  & Test       & 1M    & \makecell[l]{Python, Java, C++, C, \\ Golang, PHP, C\#, JavaScript}  \\
\midrule
\multirow{3}{*}{Task 2} 
  & Train      & 500K  & \makecell[l]{Python, Java, C++, C, Golang, \\ PHP, C\#, JavaScript, Rust} \\
  & Validation & 100K  & \makecell[l]{Python, Java, C++, C, Golang, \\ PHP, C\#, JavaScript, Rust} \\
  & Test       & 500K  & \makecell[l]{Python, Java, C++, C, Golang, \\ PHP, C\#, JavaScript, Rust} \\
\midrule
\multirow{3}{*}{Task 3} 
  & Train      & 900K  & \makecell[l]{Python, Java, C++, C, Golang, \\ PHP, C\#, JavaScript, Rust} \\
  & Validation & 200K  & \makecell[l]{Python, Java, C++, C, Golang, \\ PHP, C\#, JavaScript, Rust}  \\
  & Test       & 1M    & \makecell[l]{Python, Java, C++, C, Golang, \\ PHP, C\#, JavaScript, Rust}  \\
\bottomrule
\end{tabular}
}
\caption{\emph{AICD Bench}: Dataset statistics across tasks, splits, sample counts, and programming languages.}
\label{tab:data_splits}
\end{table}

\emph{AICD Bench} further incorporates human-written and hybrid samples from sources that were not included in \texttt{DroidCollection}. Specifically, we add 50K human-written samples in C++, C, C\#, Java, JavaScript, PHP, Go, and Rust from \emph{The Heap}~\cite{heap}. \emph{The Heap} is released as a contamination-free benchmark and does not overlap with \emph{The Stack}~\cite{stack}, which underlies datasets such as \emph{The Vault} and \emph{StarCoderData} used in \texttt{DroidCollection}.

We further augmented the \emph{AICD Bench} dataset with 100K hybrid examples from \emph{Swallow Code}~\cite{swallow}, which contains high-quality programs that were automatically generated by rewriting human-authored code using the LLaMA~3.3~70B model. The distribution of dataset splits is summarized in \Cref{tab:data_splits}.

A comparison of \texttt{DroidCollection}~\cite{droid}, \emph{CodeMirage}~\cite{codemirage}, and our dataset is shown in \Cref{tab:data_delta}. We can see that overall, \emph{AICD Bench} substantially improves both dataset scale and generator diversity, while also expanding coverage to additional programming languages and reasoning-capable models. A complete list of generator models is provided in \Cref{appx:added_models}.

\begin{table}[h!]
\centering
\resizebox{\linewidth}{!}{%
\begin{tabular}{lccc}
\toprule
\textbf{Criteria} & \texttt{DroidCollection} & \textbf{\it CodeMirage} & \textbf{\it AICD Bench} \\
\midrule
\# samples & 1.06M & 210K & \textbf{2.05M} \\
\# programming languages & 7 & \textbf{10} & 9 \\
\# models & 43 & 10 & \textbf{77}\tablefootnote{82 models if distinct quantizations are counted separately.} \\
\# model families & 10 & 6 & \textbf{11} \\
Reasoners & \xmark & \cmark & \cmark \\
\bottomrule
\end{tabular}}
\caption{Comparison of \texttt{DroidCollection}, \emph{CodeMirage}, and \emph{AICD Bench} in terms of size, languages, models, and model families.}
\label{tab:data_delta}
\end{table}

\begin{table}[tbh]
\centering
\scalebox{0.9}{
\begin{tabular}{lcc}
\toprule
\textbf{Parameter} & \textbf{Min} & \textbf{Max} \\
\midrule
AST depth & 2 & 31 \\
Maximum line length (characters) & 12 & 400 \\
Average line length (characters) & 5 & 140 \\
Number of lines of code & 6 & 300 \\
Fraction of alphanumeric characters & 0.2 & 0.75 \\
Docstring English confidence (\%) & 99 & 100 \\
\bottomrule
\end{tabular}
}
\caption{Filtering parameters used during the construction of \emph{AICD Bench}.}
\label{tab:filtering-params}
\end{table}

\subsection{Data Filtering and Quality Assurance}

We replicate \texttt{Droid}'s filtering pipeline to maintain consistency in data distribution and to prevent detectors from exploiting distributional artifacts rather than learning meaningful code patterns. The filtering process removes unparsable code, overly simple or excessively complex samples, non-code and auto-generated files, and non-English content, resulting in a clean and representative dataset. We filtered the samples according to the criteria listed in \Cref{tab:filtering-params}, which match those used in the construction of \texttt{DroidCollection} and works on Code-LM training~\cite{obscura, starcoder2, li2023starcodersourceyou, shi2025natural}. The resulting distributions (\Cref{appx:distributions}) indicate that our dataset is well structured, appropriately complex, and comparable to real-world codebases~\cite{distribution_code, djavu, commit_size}.

We performed de-duplication using the MinHash~\cite{minhash} algorithm with a similarity threshold of 0.8, applied jointly to the original and newly generated samples. This process ensures that no duplicates remain, even when samples overlap with those from \texttt{Droid}, thereby guaranteeing clean and uncontaminated data.

\subsection{Robust Binary Classification Data Subset}

Task~1 focuses on training binary detectors (human vs.\ machine) that generalize across \emph{unseen programming languages} and \emph{unseen domains}. The test set is divided into four progressively more challenging splits: (\emph{i})~seen languages (C++, Java, Python) and a seen domain (algorithmic code), (\emph{ii})~unseen languages (Golang, PHP, Rust, JavaScript, C\#, C) with a seen domain, (\emph{iii})~seen languages with unseen domains (research code and general-purpose software), and (\emph{iv})~fully unseen languages and domains. This structure enables a systematic evaluation of robustness under increasing distributional shift.

\subsection{Model Family Attribution Data Subset}

We define the following families, together with representative generators:
\begin{itemize}
    \item \textbf{DeepSeek-AI}: DeepSeek-V3~\cite{deepseekv3}, DeepSeek-R1~\cite{deepseekr1}, and DeepSeek-Coder models~\cite{deepseekcoder}.
    \item \textbf{Qwen}: Qwen3~\cite{qwen3technicalreport} and Qwen2.5~\cite{qwen2.5}.
    \item \textbf{01-ai}: Yi-Coder models~\cite{yi}.
    \item \textbf{BigCode}: StarCoder~\cite{li2023starcodersourceyou} and StarCoder2~\cite{starcoder2}.
    \item \textbf{Gemma}: Gemma3~\cite{gemma} and CodeGemma~\cite{codegemma}.
    \item \textbf{Phi}: Phi4~\cite{phi4} and Phi3~\cite{phi3}.
    \item \textbf{Meta-Llama}: Llama~\cite{llama3} and CodeLlama~\cite{codellama}.
    \item \textbf{IBM-Granite}: Granite Code~\cite{granitecode} and Granite~\cite{granite3.2}.
    \item \textbf{Mistral}: Devstral\footnote{\href{https://mistral.ai/news/devstral}{Devstral Model}}, Mixtral~\cite{mixtral}, and Mistral~\cite{mistral}.
    \item \textbf{OpenAI}: GPT-4o and GPT-4o-mini.
    \item \textbf{Gemini}: Gemini~1.5~\cite{gemini1.5} and Gemini~2.5~\cite{gemini2.5}.
\end{itemize}

We perform evaluation under two conditions: \emph{seen authors} (generators observed during training) and \emph{unseen authors} (previously unseen generators from known families). With 77 generators and an average of six models per family, this task requires fine-grained discrimination both within and across model lineages.

\subsection{Fine-Grained Human--Machine Classification Data Subset}

Task~3 distinguishes 4 categories of code: \emph{human-written}, \emph{machine-generated}, \emph{hybrid} (human-authored code rewritten or completed by an LLM), and \emph{adversarial}. The latter includes code produced using prompts or alignment strategies designed to elicit human-like outputs, including DPO-based~\cite{dpo} fine-tuning on paired human and LLM-generated solutions.

The test set contains 1M examples, with half drawn from the same sources as the training data (in-domain evaluation) and the other half drawn from \emph{The Heap} and \emph{Swallow Code} (out-of-domain evaluation), enabling a direct assessment of generalization under domain shift.

\section{Experiments and Results}

\subsection{Experimental Setup}
To assess the utility of \emph{AICD Bench}, we train a suite of encoder-based classifiers on its three tasks. Unless stated otherwise, all models are trained for 3 epochs with a batch size of 64 and an input window of 512 tokens, following the configuration that yields the best performance in \texttt{Droid}. Importantly, our goal is not to obtain state-of-the-art results; rather, these experiments aim to demonstrate the benchmark’s applicability and to confirm that it provides a meaningful and reliable evaluation.

We experimented with the following encoders: CodeBERT~\cite{codebert}, CodeT5+~\cite{codet5+}, UnixCoder~\cite{unixcoder}, ModernBERT~\cite{modernbert}, RoBERTa~\cite{roberta}, and DeBERTa-v3 (hereafter, DeBERTa)~\cite{deberta}. We selected them due to their widespread use in AI-generated content detection: CodeBERT is used by \citet{gptsniffer}, CodeT5+ and UnixCoder by \citet{gptsensor} and \citet{codet}, ModernBERT by \citet{droid}, and RoBERTa by \citet{m4}.

We use Macro-F1 as the primary evaluation metric. Since the datasets and the tasks are class-imbalanced, we focus on reliable performance across all classes rather than optimizing for the majority label. Macro-F1 computes the F1 score independently for each class and then averages across classes, ensuring that minority-class performance contributes equally to the overall score. Compared to accuracy (often dominated by frequent classes) and Micro-F1 (which aggregates over all instances and thus favors majority classes), Macro-F1 provides a fairer and more informative assessment in our setting.

In addition to encoder-based models, we train classical baselines, including Logistic Regression, SVM, and CatBoost~\cite{catboost}. For each, we evaluate three feature representations: (\emph{i})~TF-IDF, (\emph{ii})~AST-based features, which have been shown to be useful for AI-generated code detection~\cite{codet,droid,whodunit}, and (\emph{iii})~their combination. 

For TF-IDF, we use the full source text and tokenize it using whitespace delimiters, i.e.,\ any sequence of spaces, tabs, or line breaks, without applying lowercasing, stop-word removal, or punctuation/identifier normalization. From this token stream, we construct uni-, bi-, and tri-gram vocabularies and compute TF-IDF features over the training corpus, starting from 5{,}000 n-gram features and reducing dimensionality to 500 using truncated singular value decomposition (SVD; \citealp{svd}). For AST-based features, we parse each program using Tree-Sitter\footnote{\href{https://tree-sitter.github.io/tree-sitter/}{https://tree-sitter.github.io/tree-sitter/}} and extract structural signals, including overall AST depth, counts of node/construct types (\emph{e.g.}, numbers of \texttt{if} statements, loops, and function definitions), layout/style indicators (\emph{e.g.}, empty-line density), and code-complexity proxies. Feature dimensionality is data-driven and equals the number of distinct AST-derived attributes observed in training, yielding 543 features for the Robust Binary Classification task and 1{,}030 for the remaining tasks (which involve more programming languages).

We also run zero-shot Fast-De\-tectGPT~\cite{fast}, a faster version of De\-tectGPT~\cite{detectgpt} that uses human-machine differences in token-choice likelihood and has shown strong performance detecting AI text and code~\cite{codet}. Since it is best suited to binary decisions, we evaluate it on Task 1 and on a binarized version of Task 3 (human vs.\ all other classes). Finally, we evaluate LLM performance on proposed tasks; see \Cref{appx:llms} for details.

\begin{table}[!t]
\centering
\tiny
\resizebox{0.42\textwidth}{!}{%
\begin{tabular}{lccc}
\toprule
\textbf{Model} & \textbf{Task 1} & \textbf{Task 2} & \textbf{Task 3}\\
\midrule
Random\tablefootnote{\blue{Scores differ from 1/\#classes due to class imbalance.}} & \underline{45.73} & 5.69 & 20.34 \\
Majority & \underline{43.83} & 5.43 & 20.05 \\
Fast-DetectGPT & 44.99 & - & 50.03\tablefootnote{Binarized.} \\
\midrule
CB$_{\text{TF-IDF}}$ & 29.40 & 5.47 & 22.45 \\
CB$_{\text{AST}}$ & 18.02 & 7.78 & 14.59 \\
CB$_{\text{AST \& TF-IDF}}$ & 18.02 & 0.12 & 3.09 \\
\midrule
SVM$_{\text{TF-IDF}}$ & \textbf{43.05} & 5.44 & 21.28 \\
SVM$_{\text{AST}}$ & 18.02 & 5.99 & 3.43 \\
SVM$_{\text{AST \& TF-IDF}}$ & 18.02 & 1.38 & 5.82 \\
\midrule
LR$_{\text{TF-IDF}}$ & 37.10 & 5.45 & 20.64 \\
LR$_{\text{AST}}$ & 18.02 & 5.44 & 3.18 \\
LR$_{\text{AST \& TF-IDF}}$ & 18.02 & 1.09 & 5.23 \\
\midrule
CodeBERT & 28.64 & 23.71 & 55.60 \\
CodeT5+ & 28.08 & 5.43 & 52.76 \\
ModernBERT & 30.61 & \textbf{32.84} & \textbf{61.65} \\
RoBERTa & 31.88 & 17.52 & 52.05 \\
UnixCoder & 25.51 & 26.64 & 54.21 \\
DeBERTa & 34.13 & 12.08 & 54.65 \\
\bottomrule
\end{tabular}
}
\caption{Macro F1-score of baselines across all tasks. Best model performance per task is \textbf{bolded}, while dummy baselines (random or majority) are \underline{underlined} where they outperform trained models. (Logistic Regression - LR; CatBoost - CB. Task 1 corresponds to the Robust Binary Classification task, Task 2 corresponds to the Model Family Attribution task, and Task 3 corresponds to the Fine-Grained Human-Machine Classification task.)}
\label{tab:results}
\end{table}

\subsection{Evaluation Results}
\Cref{tab:results} shows that deep learning models (CodeBERT, CodeT5+, ModernBERT, RoBERTa, UniXCoder, and DeBERTa) outperform classical approaches. This suggests that neural encoders' contextual representations provide more informative signals for identifying AI-generated code than structural (AST-based) or statistical (TF-IDF-based) features. Among neural models, ModernBERT achieves the strongest performance on Tasks 2 and 3, consistent with findings reported by \citet{droid}. Zero-shot Fast-DetectGPT outperforms trained models in Task 1 but still falls short of the random baseline, highlighting the difficulty of robust binary detection under distribution shift.

On Task 3, even when evaluated in a binarized setting, Fast-DetectGPT lags behind transformer-based detectors, indicating that hybrid and adversarial samples diverge substantially from what this detector characterizes as machine-generated.

\subsubsection{Task 1: Robust Binary Classification}
\label{sec:task1}
In this task, classical models achieve unexpectedly strong performance (see \Cref{tab:results}): SVM and Logistic Regression with TF-IDF features outperform all deep learning models. Because the task primarily probes out-of-distribution generalization, this behavior is consistent with classical learning theory, where increased model complexity can exacerbate overfitting to spurious correlations~\cite{rohlfs_generalization}. At the same time, both random and majority-class baselines achieve even higher scores, highlighting a severe train–test distribution shift that limits reliable generalization.

Further analysis indicates that language-independent cues, such as variable naming patterns, play a central role in distinguishing human-written from AI-generated code (see \Cref{appx:svm}), which helps explain why TF-IDF features are particularly effective in this setting.

\begin{table}[h]
\centering
\resizebox{\linewidth}{!}{%
\begin{tabular}{lcc}
\toprule
\textbf{Setting} & \textbf{Macro F1} & \textbf{STD ($\pm$)} \\
\midrule
Seen domain, seen language     & 0.63 & 0.30 \\
Seen domain, unseen language   & 0.42 & 0.12 \\
Unseen domain, unseen language & 0.21 & 0.07 \\
Unseen domain, seen language   & 0.20 & 0.08 \\
\bottomrule
\end{tabular}%
}
\caption{\textbf{Task 1 (Robust Binary Classification):} experiments across different domain/language settings. We report Macro F1, averaged over detectors.}
\label{tab:task_1_combined}
\end{table}

We further analyze model performance across four input scenarios: (\emph{i})~seen language and seen domain, (\emph{ii})~unseen language in a seen domain, (\emph{iii})~seen language in an unseen domain, and (\emph{iv})~unseen language and an unseen domain. As shown in \Cref{tab:task_1_combined}, domain shift emerges as the dominant source of error: models perform nearly as poorly when only the domain is unseen as when both the domain and the language are unseen. This behavior is expected, as coding style and conventions vary substantially across domains; for instance, research code often contains extensive explanatory comments, whereas algorithmic problem solutions are typically concise and minimally annotated.

As shown in \Cref{appx:task_1_performance}, the AST features decrease the performance of the model when paired with TF-IDF. We believe that it is reasonable, since the structure of the unseen data is very different from the training data, so the AST features cannot be used as a robust representation. Statistical features from TF-IDF, at the same time, can handle not only the syntactical, but also some stylistical (naming conventions, etc.) properties of the code, which can be used to identify LLM-generated code.

\subsubsection{Task 2: Model Family Attribution}
\label{sec:task2}

As shown in Tables~\ref{tab:results} and \ref{tab:task_2_combined}, this task is the most challenging among all three tasks, exhibiting the lowest peak and average performance across models. This difficulty is largely attributable to the increased number of classes: while Robust Binary Classification involves only two labels and Fine-Grained Human-Machine Classification involves four, model family attribution requires discrimination among 12 classes.

A pronounced performance gap is also observed between in-domain and out-of-domain generators. As shown in \Cref{tab:task_2_combined}, attributing previously unseen generators to their corresponding families proves particularly difficult, suggesting substantial variation within model families and limited transferability of learned stylistic signals. 

\begin{table}[tbh]
\centering
\resizebox{\linewidth}{!}{%
\begin{tabular}{lcc}
\toprule
\textbf{Category} & \textbf{Macro F1} & \textbf{STD ($\pm$)} \\
\midrule
In-domain Generator     & 0.149 & 0.154 \\
Out-of-Domain Generator & 0.046 & 0.070 \\
\bottomrule
\end{tabular}%
}
\caption{\textbf{Task 2 (Model Family Attribution):} average Macro-F1 across models by generator type.}
\label{tab:task_2_combined}
\end{table}

\begin{figure}[h]
    \centering
    \includegraphics[width=1.0\linewidth]{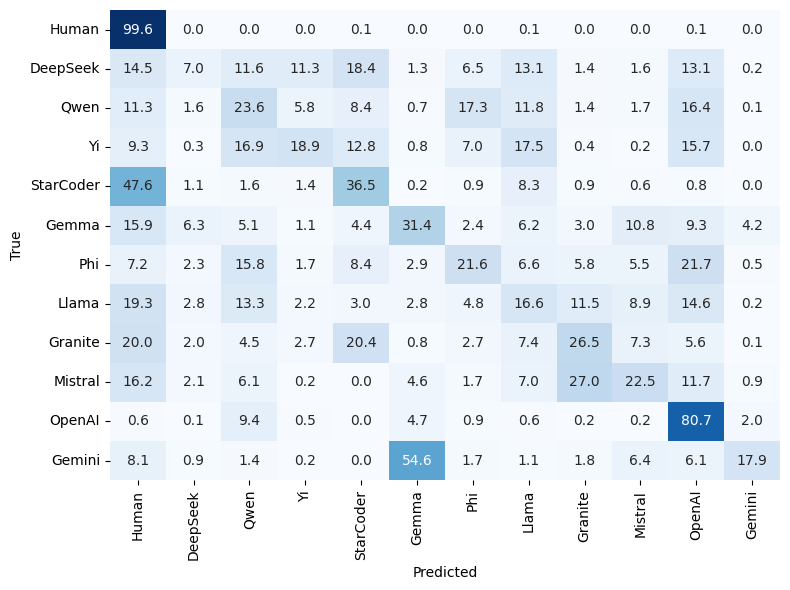}
    \caption{\textbf{Task 2 (Model Family Attribution):} ModernBERT confusion matrix. The values are row-normalized percentages, showing the proportion of each true class assigned to each predicted class.}
    \label{fig:task_2_modernbert}
\end{figure}

Moreover, the standard deviation of performance exceeds the mean, which together with \Cref{fig:task_2_modelwise} indicates high variability and instability across classification models.

\Cref{fig:task_2_modernbert} further illustrates these challenges. The best-performing model, ModernBERT, almost perfectly identifies human-written code and OpenAI-generated samples, which can be attributed to the low intra-family variability of these classes: they comprise only one (human) and two (GPT-4o and GPT-4o-mini) generators, respectively. In contrast, Gemini samples are frequently misclassified as Gemma, likely reflecting shared design and training characteristics, as both model families are developed by Google. We also observe that StarCoder outputs are often misclassified as human-written code, which is consistent with the fact that the StarCoder family includes multiple base models and is trained predominantly on large volumes of open-source human-authored code. This training regime may lead to generations that closely resemble natural coding patterns in terms of structure, naming conventions, and style. A detailed, model-wise analysis for this task is provided in \Cref{appx:task_2_performance}.

\begin{table}[tbh]
\centering
\resizebox{0.9\linewidth}{!}{%
\begin{tabular}{lcc}
\toprule
\textbf{Category} & \textbf{Macro F1} & \textbf{STD ($\pm$)} \\
\midrule
In-domain     & 0.366 & 0.302 \\
Out-of-Domain & 0.119 & 0.094 \\
\bottomrule
\end{tabular}%
}
\caption{\textbf{Task 3 (Fine-Grained Human-Machine
Classification)}: average Macro-F1 across classifiers by generator type with std.}
\label{tab:task_3_combined}
\end{table}

\begin{figure}[h]
    \centering
    \includegraphics[width=\linewidth]{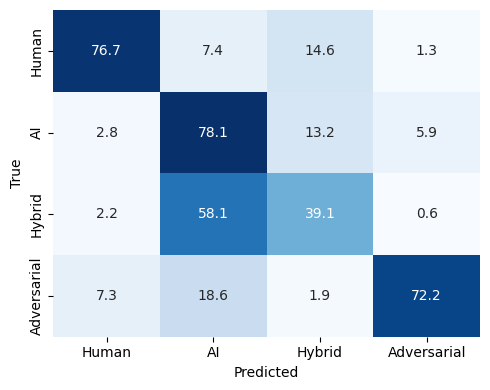}
    \caption{\textbf{Task 3 (Fine-Grained Human-Machine Classification)}: ModernBERT confusion matrix. The values are row-normalized percentages, showing the proportion of each true class assigned to each predicted class.}
    \label{fig:task_3_modernbert}
\end{figure}

\subsubsection{Task 3: Fine-Grained Human-Machine Classification}
\label{sec:task3}
As shown in Tables~\ref{tab:results} and \ref{tab:task_3_combined}, in this multi-class setting, deep learning models consistently outperform classical approaches. Among the classical baselines, the strongest results are obtained using TF-IDF features alone, which we attribute to the sparsity of the AST-based representation: out of 1{,}030 AST features, only 75\% are non-zero. This sparsity limits the effectiveness of structural cues in distinguishing subtle authorship differences across fine-grained classes.

A substantial performance gap emerges when comparing in-domain and out-of-domain evaluation, particularly for samples drawn from \emph{Swallow Code} and \emph{The Heap}. This gap indicates that models struggle to generalize across data distributions that differ from those observed during training (see \Cref{tab:task_3_combined}). Notably, this degradation persists even for strong encoder models, suggesting that fine-grained detection is especially sensitive to domain shift. A more detailed dataset-level analysis is provided in \Cref{appx:task_3_performance} for completeness and further insights.

An examination of the predictions from the best-performing model, ModernBERT, reveals that most errors occur on out-of-domain examples, especially \emph{Swallow Code} instances labeled as \emph{hybrid}. These examples are predominantly misclassified as fully AI-generated (see \Cref{fig:task_3_modernbert}). In addition, adversarial examples are frequently misclassified as either AI-generated or human-written, which is expected given that such code is deliberately crafted to mimic human authorship.

Overall, although this task is comparatively easier than the other two, as is reflected in generally higher scores across models, the best achieved performance (a Macro-F1 of 61.65) remains well below practical requirements. This underscores the continued difficulty of fine-grained human–machine discrimination and highlights the challenge posed by \emph{AICD Bench}.

\begin{table}[t]
\centering
\tiny
\resizebox{\linewidth}{!}{%
\begin{tabular}{lccc}
\toprule
\textbf{Model} & \textbf{Task 1} & \textbf{Task 2} & \textbf{Task 3}\\
\midrule
Gemini$_{zs}$ & 57.15 & \textbf{5.43} & 25.95\\
Gemma$_{zs}$ & 58.45 & 5.37 & 22.21\\
\midrule
Gemini$_{cot}$ & \textbf{62.31} & 5.54 & \textbf{28.14} \\
Gemma$_{cot}$ & 54.94 & 5.37 & 22.23\\
\bottomrule
\end{tabular}
}
\caption{\textbf{Zero-shot LLM experiments across all three tasks:} Macro F1-scores of LLMs with simple zero-shot prompting (zs) and chain-of-thought prompting (cot).}
\label{tab:llm_results}
\end{table}

\subsubsection{Zero-Shot LLM Experiments}
\label{appx:llms}

Next, we evaluate the ability of large language models to perform the tasks defined in \emph{AICD Bench}, in a zero-shot setting, using two representative LLMs: Gemini-2.5-Flash and Gemma3-27B. To ensure cost-efficient yet representative evaluation, we sample 5\% of the data per task, stratified by label, domain, and programming language. We consider two evaluation settings: zero-shot classification and classification with chain-of-thought (CoT) prompting. The prompts used for both settings are provided in \Cref{appx:prompts}.

As shown in \Cref{tab:llm_results}, zero-shot CoT prompting consistently improves the performance across tasks. In particular, Gemini with CoT prompting achieves the highest score on Task 1 among all evaluated models. However, Task 2 remains highly challenging for LLMs, and the performance on Task 3 is substantially lower than that of trained detection models.

\subsection{Error Analysis}
\label{sec:errors}

We analyze errors by examining the cases where all evaluated models fail; representative examples are shown in \Cref{appx:errors}.

For Task~1, we observe a clear asymmetric failure pattern: all models misclassify certain human-written samples as AI-generated, while no AI-generated samples are misclassified by all models. This suggests that detectors effectively identify AI-generated code, likely due to shared stylistic artifacts, but struggle with the broader syntactic, structural, and stylistic diversity of human-written code. Manual inspection reveals no consistent LLM-like traits (e.g., boilerplate, verbose comments, repetitive patterns, or unusual naming), indicating that these are genuine human examples flagged because they fall outside the ``typical'' human patterns seen during training.

For Task~2, we observe no consistent error patterns across models. In contrast, for Task~3, all models fail on hybrid and human-written samples from \emph{Swallow Code} and \emph{The Heap}, which are absent from the training data. This further highlights the strong sensitivity of detectors to domain shifts, particularly in fine-grained classification settings.

We also apply SHAP~\cite{shap} to analyze which input features most strongly influence the model predictions; illustrative examples are shown in \Cref{appx:shap}. For Task~1, deep models exhibit a pronounced bias toward competitive-programming artifacts: surface-level templates (e.g., \texttt{\#include}, \texttt{\#define}, \texttt{class Solution}, loop scaffolding, and typed signatures) are overemphasized relative to semantic content, while explanatory or demonstration tokens (e.g., \texttt{the answer would}, \texttt{TreeNode}, \texttt{sum}, \texttt{summary}, \texttt{//}) often trigger incorrect AI predictions under domain shift.

In contrast, for Task~3 especially for hybrid samples the most informative cues arise from \emph{mixed contexts}, such as non-algorithmic headers interleaved with code, shell/URL/path fragments near code tokens, and documentation-like text embedded within code (docstrings, small unit tests, usage examples, repeated imports, and high-level task verbs such as \texttt{deploy}, \texttt{retrieve}, or \texttt{server}).

Features that push predictions away from the hybrid class include runnable scaffolding (e.g.,~\texttt{if \_\_name\_\_ == \_\_main\_\_}), dense CLI/IO patterns, explicit error handling (\texttt{try/except}), and human-written license headers. Overall, codestyle mixing is the dominant signal for identifying hybrid code.

\section{Conclusion and Future Work}

We presented \emph{AICD Bench}, the largest and most comprehensive benchmark for AI-generated code detection to date. It comprises 2M code samples from 77 generators, including strong reasoning-oriented models, across 9 programming languages, and introduces three complementary tasks: robust binary classification under distribution shifts, model family attribution, and fine-grained human–machine classification. By jointly emphasizing scale, diversity, and realistic evaluation settings, \emph{AICD Bench} substantially extends the scope of prior benchmarks.

Extensive evaluations with both neural and classical baselines demonstrate that these tasks remain far from solved. Current detectors struggle to generalize across programming languages, domains, and generator families, and perform particularly poorly on hybrid and adversarial code. While ModernBERT achieves the strongest overall performance, leading on two of the three tasks, even the best-performing models fall well short of practical requirements. Notably, a simple SVM outperforms deep learning models on robust binary classification, yet still performs below random guessing, highlighting severe generalization challenges and the urgent need for new detection paradigms.

Beyond exposing these limitations, \emph{AICD Bench} provides a standardized and extensible evaluation framework with unified tasks, splits, and protocols, enabling reproducible comparison and systematic progress. By moving beyond oversimplified in-distribution binary detection, it reorients the field toward realistic, deployment-driven challenges.

In future work, we plan to investigate adversarial and domain-adaptive training strategies aimed at improving robustness under distribution shift, such as adversarial data augmentation, domain-invariant representation learning, and curriculum-based adaptation across programming languages and domains. We also intend to develop meta-models that explicitly promote generalization across languages, domains, and generator families, for example by learning shared detection priors or dynamically combining task-specific detectors. Finally, we aim to build an automated pipeline to continuously expand the benchmark with newly released LLMs, emerging programming languages, and diverse sources of human-written code, ensuring that \emph{AICD Bench} remains representative of evolving real-world coding practices.

\section*{Limitations}

\paragraph{Potential Data Contamination.}
As with many public benchmarks, there is a risk that \emph{AICD Bench} may become saturated over time as models are increasingly tuned to its specific distributions. This is particularly problematic given that the benchmark is intended to assess out-of-distribution robustness. To mitigate this risk, we plan to introduce a private evaluation split with hidden labels, where submissions will be evaluated through an online platform rather than via local testing.

\paragraph{Reliance on DroidCollection.}
Our benchmark builds substantially on \texttt{DroidCollection}. Despite incorporating additional models, two supplementary datasets, and new programming languages, it remains influenced by the underlying distribution of that resource. Consequently, (\emph{i})~none of the \texttt{Droid} generators can be directly evaluated on \emph{AICD Bench}, and (\emph{ii})~the observed performance trends may still reflect distributional skews inherited from \texttt{DroidCollection} rather than those of AI-generated and human-written code encountered in real-world settings. We plan to address this limitation by iteratively updating the benchmark with newer and more diverse data sources, thereby reducing reliance on any single dataset.

\paragraph{Constraints on Code Diversity.}
Although \emph{AICD Bench} aims to approximate real-world coding scenarios, its coverage remains constrained along two dimensions. First, the dataset is filtered using parameters such as AST depth, line length, and code size, which ensures clean samples but excludes highly verbose, overly complex, extremely short, or poorly structured code commonly found in practice. As a result, the benchmark may not fully capture the syntactic, structural, and stylistic variability of human-written code. Second, the benchmark currently spans nine widely used programming languages (C++, C, C\#, Go, Java, JavaScript, PHP, Python, and Rust), which, while representative of many real-world applications, form a finite set and do not test generalization beyond this scope.

To mitigate these limitations, future iterations of \emph{AICD Bench} will (\emph{i})~relax filtering thresholds to include more atypical and low-quality human-written code, and (\emph{ii})~expand language coverage to less mainstream or domain-specific languages (e.g.,~Swift, Kotlin, MATLAB), enabling evaluation under broader and more challenging distributional shifts.

\section*{Ethical Statement}
\label{sec:ethics}

\paragraph{Data Collection and Privacy.}
\emph{AICD Bench} is constructed from publicly available research corpora and model outputs generated via documented interfaces. We comply with platform terms of service and respect upstream licenses. Human-written code is sourced from published research datasets; no private repositories, paywalled content, or sensitive personal identifiers are included. Where required, we preserve original attributions and sufficient metadata for license compliance.

Our work aims to promote transparency in AI-assisted coding and support applications such as plagiarism detection, compliance, and provenance auditing. To mitigate misuse risks (\emph{e.g.}, detector evasion), we do not release exploit-oriented prompt details and clearly document the limitations of current detectors.

\paragraph{Bias.}
Both human- and LLM-authored code may reflect biases arising from data availability, platform popularity, community conventions, and training corpora. As a result, \emph{AICD Bench} may inherit distributional skews (\emph{e.g.}, language, domain, or style imbalances) that affect external validity. We mitigate these risks through diverse sampling across platforms, languages, and generator families, though perfect representativeness cannot be guaranteed. Future releases will include more detailed bias analyses and refined sampling strategies.

\paragraph{Risk of Misuse.}
Although \emph{AICD Bench} strengthens detection research, it could also be misused to develop evasion strategies. To reduce this risk, we avoid releasing adversarial prompts and emphasize that the benchmark is for research use only.

\paragraph{Broader Impact.}
By framing detection as a multitask, realism-driven benchmark, we aim to move the field beyond oversimplified binary detection toward practical robustness. Despite its limitations, \emph{AICD Bench} provides a foundation for more transparent, accountable, and trustworthy AI-assisted programming.

\section*{Acknowledgments}

\blue{At TU Darmstadt, this research has been supported by the German Federal Ministry of Research, Technology and Space and the Hessian Ministry of Higher Education, Research, Science and the Arts within their joint support of the National Research Center for Applied Cybersecurity ATHENE.}

\bibliography{custom}

\vfill\eject


\clearpage

\appendix

\textbf{\large{Appendix}}

\section{Data Statement}
\label{sec:data_statement}

\paragraph{A.1 General Information}

\paragraph{Dataset Title} \textbf{AICD Bench}

\paragraph{Dataset Version} 1.0 (September 2025)

\paragraph{Data Statement Version} 1.0 (September 2025)

\paragraph{A.2 Executive Summary}

\emph{AICD Bench} is designed for a rigorous and standalone evaluation of systems that distinguish human-written code from machine-generated, hybrid, and adversarial code across multiple programming languages, code generators, and domains. It compiles a large and diverse set of code snippets drawn from widely used programming platforms and contemporary LLM code generators, ensuring broad language and domain coverage.



\textbf{Intended Use:} \emph{AICD Bench} is intended exclusively for research, particularly for developing and evaluating models that detect machine-generated code. Researchers can analyze how programming languages, generation models, and application domains affect detection accuracy and robustness. The dataset aims to support improved automated code assessment, ethical use, and accountability in software engineering.

\textbf{Usage Restrictions:} The dataset is provided solely for academic and research purposes. Commercial use is prohibited without prior written consent from the dataset creators. Users must follow ethical guidelines and ensure that the findings do not violate privacy, intellectual property, or other legal constraints. Redistribution is not permitted without authorization from the dataset custodians.

\textbf{Source:} We open-source all the splits of \emph{AICD-Bench} on Hugging Face.\footnote{\url{https://huggingface.co/AICD-bench}}

\paragraph{License}
\textbf{Creative Commons Attribution--NonCommercial 4.0 International (CC BY--NC 4.0)}. We will release \textbf{AICD Bench} under CC BY--NC 4.0. You may share and adapt the dataset for \emph{non-commercial} research with appropriate attribution; no additional restrictions may be applied. For any commercial use, prior written permission from the authors is required. See the full license at \href{https://creativecommons.org/licenses/by-nc/4.0/}{https://creativecommons.org/licenses/by-nc/4.0/}. The dataset is provided ``as is,'' without warranties of any kind.



\section{Added Models}
\label{appx:added_models}
We enlarge the \texttt{DroidCollection}~\cite{droid} with additional models for nearly every family of models. \Cref{tab:models} indicates which models were taken from the original \texttt{DroidCollection}, and which were introduced in \emph{AICD Bench}.

\begin{table}[!tbp]
\centering
\scalebox{0.6}{
\begin{tabular}{ll}
\toprule
\textbf{\texttt{Model Family}} & \textbf{\texttt{Model}} \\
\midrule

\multirow{4}{*}{\textbf{Yi}} & Yi-Coder-9B \\
& Yi-Coder-9B-Chat \\
& Yi-Coder-1.5B-Chat \\
& Yi-Coder-1.5B   \\
\midrule

\multirow{2}{*}{\textbf{GPT}} & GPT-4o-mini \\
& GPT-4o  \\
\midrule

\multirow{11}{*}{\textbf{Qwen}} & Qwen2.5-Coder-7B \\
& Qwen2.5-Coder-7B-Instruct \\
& Qwen2.5-Coder-1.5B-Instruct \\
& Qwen2.5-Coder-32B-Instruct \\
& Qwen2.5-72B-Instruct \\
& Qwen2.5-Coder-1.5B \\
& Qwen2.5-Coder-14B-Instruct   \\
& \textbf{\textit{QwQ-32B}} \\
& \textbf{Qwen3-14B} \\
& \textbf{Qwen3-32B} \\
& \textbf{\textit{Qwen3-235B-A22B}} \\
& \textbf{\textit{Qwen3-30B-A3B}} \\
\midrule

\multirow{7}{*}{\textbf{Gemma}} & codegemma-7b-it \\
& codegemma-7b \\
& codegemma-2b \\
& \textbf{gemma-3-27b-it} \\
& \textbf{gemma-3n-e4b-it} \\
& \textbf{gemma-3-12b-it} \\
& \textbf{gemma-3-4b-it} \\
\midrule


\multirow{6}{*}{\textbf{Deepseek}} & deepseek-coder-6.7b-instruct \\
& deepseek-coder-6.7b-base \\
& deepseek-coder-1.3b-instruct \\
& deepseek-coder-1.3b-base \\
& \textbf{\textit{DeepSeek-R1}} \\
& \textbf{DeepSeek-V3-0324} \\
\midrule

\multirow{8}{*}{\textbf{Granite}} & granite-8b-code-instruct-4k \\
& granite-8b-code-base-4k \\
& \textbf{granite-3.2-2b-instruct} \\
& \textbf{granite-3.3-8b-base} \\
& \textbf{granite-3.3-8b-instruct} \\
& \textbf{granite-34b-code-instruct-8k} \\
& \textbf{granite-3b-code-base-128k} \\
& \textbf{granite-3b-code-instruct-128k} \\
\midrule

\multirow{15}{*}{\textbf{Llama}} & Llama-3.1-8B-Instruct \\
& Llama-3.2-3B \\
& Llama-3.1-70B-Instruct \\
& Llama-3.3-70B-Instruct \\
& Llama-3.3-70B-Instruct-Turbo \\
& Llama-3.2-1B \\
& Llama-3.1-8B \\
& CodeLlama-70b-Instruct-hf \\
& CodeLlama-34b-Instruct-hf \\
& CodeLlama-7b-hf \\
& \textbf{Llama-3.2-11B-Vision-Instruct} \\
& \textbf{Llama-3.2-90B-Vision-Instruct} \\
& \textbf{Llama-4-Maverick-17B-128E-Instruct-FP8} \\
& \textbf{Llama-4-Scout-17B-16E-Instruct} \\
& \textbf{Meta-Llama-3.1-405B-Instruct} \\
\midrule

\multirow{7}{*}{\textbf{Phi}} & Phi-3-small-8k-instruct \\
& Phi-3-mini-4k-instruct \\
& phi-4 \\
& Phi-3-medium-4k-instruct \\
& phi-2 \\
& Phi-3.5-mini-instruct \\
& \textbf{Phi-4-multimodal-instruct} \\
\midrule

\multirow{6}{*}{\textbf{Mistral}} & Mistral-Small-24B-Instruct-2501 \\
& \textbf{Devstral-Small-2505} \\
& \textbf{Mistral-7B-Instruct-v0.3} \\
& \textbf{Mistral-Nemo-Instruct-2407} \\
& \textbf{Mixtral-8x7B-Instruct-v0.1} \\
\midrule

\multirow{6}{*}{\textbf{BigCode}} & starcoder2-15B \\
& starcoder \\
& starcoder2-7b \\
& starcoder2-3b \\
& \textbf{starcoderbase-1b} \\
& \textbf{starcoderbase-3b} \\
\midrule

\multirow{5}{*}{\textbf{Gemini}} & \textbf{gemini-1.5-flash} \\
& \textbf{gemini-1.5-flash-8b} \\
& \textbf{gemini-2.0-flash} \\
& \textbf{gemini-2.0-flash-lite} \\
& \textbf{\textit{gemini-2.5-flash-preview-05-20}} \\
\midrule

\end{tabular}
}
\caption{Model families and their selected models used in \emph{AICD Bench}. \textbf{Bold} entries are models that are not used in \texttt{DroidCollection}. \textit{Italic} entries are reasoning models.}
\label{tab:models}
\end{table}
\section{Data Distribution}
\label{appx:distributions}
\begin{figure*}
    \centering
    \includegraphics[width=\linewidth]{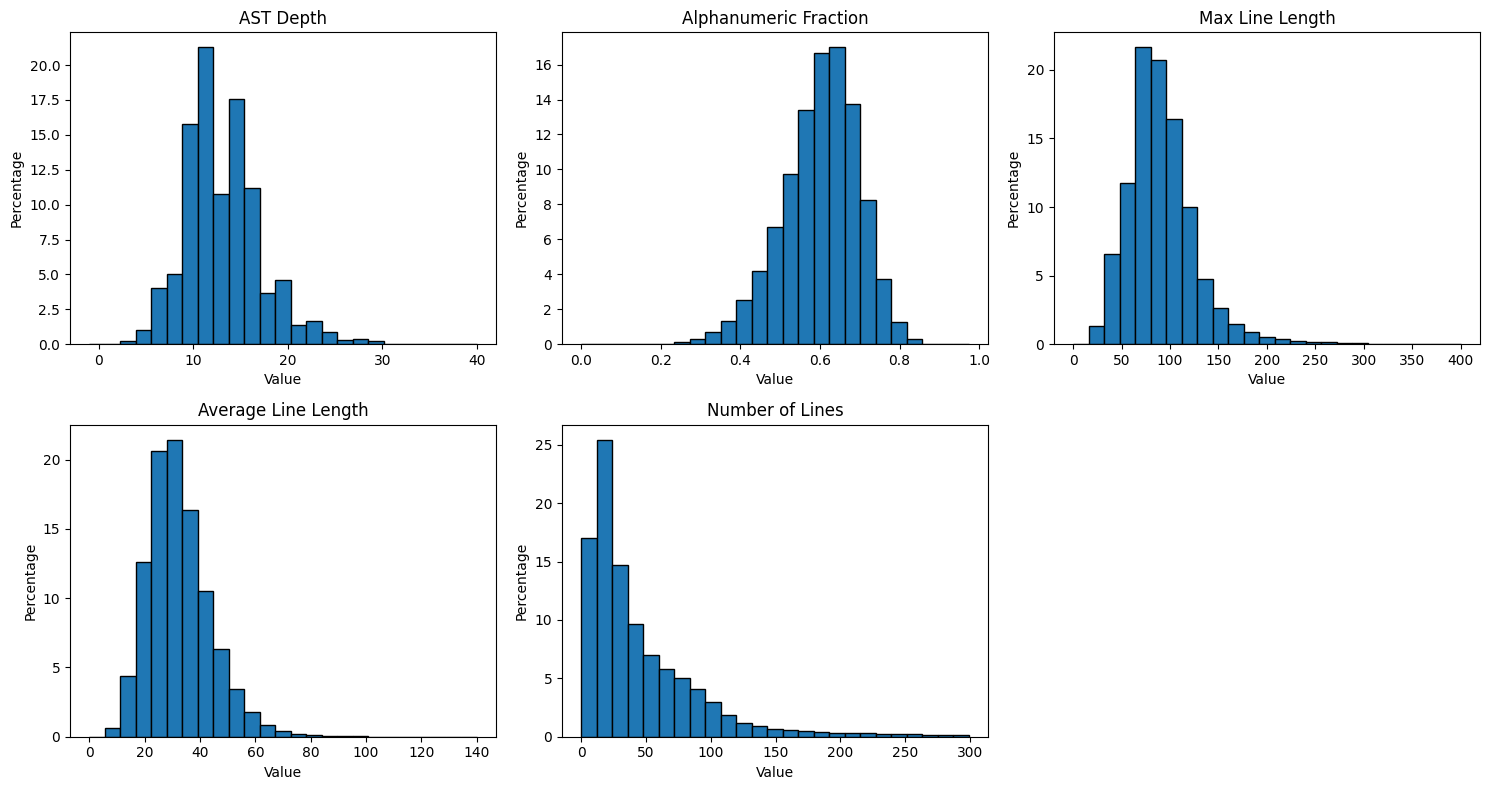}
    \caption{Distribution of code properties.}
    \label{fig:data_distribution}
\end{figure*}

\Cref{fig:data_distribution} shows that the code in our dataset is consistent with healthy coding practices of real-world projects. Namely, AST depth is moderately concentrated (most of the values are between 10 and 15), indicating a prevalence of structured but not overly nested control flow; the alphanumeric fraction peaks at 0.65 with a small standard deviation, reflecting clear, meaningful identifiers in the code, not noise, logs or obfuscated files; average line length clusters tightly between 20-60 characters, while max line length exhibits a right-skewed tail extending beyond 200 characters reflecting adherence to the best practices of code structuring; and file size is heavily skewed toward short snippets (<100 lines), with a median of nearly 40 lines, indicative of fine-grained code units that would fit into the context window of most of the classifiers.

\subsection{Prompt Templates}
\label{appx:prompts}

\lstset{
  basicstyle=\ttfamily\small, 
  lineskip=-1pt,              
  aboveskip=5pt, belowskip=5pt, 
  breakatwhitespace=true, 
}
\begin{tcolorbox}[
  colback=green!15,
  colframe=green!20,
  title=Task 1 zero-shot prompt,
  fonttitle=\bfseries,
  coltitle=black,
  boxrule=0.3pt,
  arc=1mm,
  enhanced,
  breakable
]

\begin{lstlisting}
Reply with exactly one word
Human or AI.
Given this code,
identify its origin
{code}
\end{lstlisting}
\end{tcolorbox}
\begin{tcolorbox}[
  colback=green!15,
  colframe=green!20,
  title=Task 2 zero-shot prompt,
  fonttitle=\bfseries,
  coltitle=black,
  boxrule=0.3pt,
  arc=1mm,
  enhanced,
  breakable
]

\begin{lstlisting}
Reply with exactly one word from 
this list: 
human, deepseek-ai,
Qwen,01-ai,
bigcode,Gemma,
Phi,meta-llama,ibm-granite,mistralai,
OpenAI,Gemini.
Given this code,
identify its origin
{code}
\end{lstlisting}
\end{tcolorbox}

\begin{tcolorbox}[
  colback=green!15,
  colframe=green!20,
  title=Task 3 zero-shot prompt,
  fonttitle=\bfseries,
  coltitle=black,
  boxrule=0.3pt,
  arc=1mm,
  enhanced,
  breakable
]
\begin{lstlisting}
Reply with exactly one word
Human, AI, Hybrid or Adversarial.
Given this code, 
was it written by Human, AI,
in Hybrid collaboration or by an
Adversarial model which 
tried to fool the detector mimicing 
human? {code}
\end{lstlisting}
\end{tcolorbox}

For CoT prompts, we summarized the key code features per class using Gemini-2.5-Flash.

\begin{tcolorbox}[
  colback=green!15,
  colframe=green!20,
  title=Task 1 CoT prompt,
  fonttitle=\bfseries,
  coltitle=black,
  boxrule=0.3pt,
  arc=1mm,
  enhanced,
  breakable
]
\begin{lstlisting}[breaklines=true],
You are a code-origin classifier.
Infer whether a given code snippet
was written by a Human or an AI assistant.
Reason silently and do NOT reveal
your reasoning.
Output exactly one word: Human or AI.

When deciding, consider
(but do not list) signals such as:
- Naming: verbosity/descriptiveness,
camelCase/snake_case consistency, oddly generic names.
- Comments/docstrings: density, uniform template-like phrasing, section headers, per-function summaries.
- Type hints & annotations: pervasive, perfectly consistent typing vs. ad-hoc/mixed usage.
- Structure/style: consistent formatter fingerprints (Black/Prettier), tidy imports, exhaustive edge-cases, spotless spacing;
  versus idiosyncratic style, inconsistencies, TODO/WIP markers.
- Boilerplate/templates: license blocks, README headers, instructional comments, scaffolded regions, foo/bar placeholders,
  repeated auto-generated patterns.
- API/library usage: unused imports, verbose defensive code, redundant checks, generic logging.
- Testing/examples: synthetic data, contrived examples, pedantic error messages.
- AI artifacts: phrases like "This function...", tutorial-like narration, or code mirroring common LLM examples.
- Human artifacts: partial implementations, inline hacks, commented-out experiments, domain-specific shortcuts, mismatched styles, quick fixes.

If uncertain, choose the more likely origin using the balance of signals.
Respond with a single word only.

Code:
```{code}```
\end{lstlisting}
\end{tcolorbox}

\begin{tcolorbox}[
  colback=green!15,
  colframe=green!20,
  title=Task 2 CoT prompt,
  fonttitle=\bfseries,
  coltitle=black,
  boxrule=0.3pt,
  arc=1mm,
  enhanced,
  breakable
]
\begin{lstlisting}
You are a code-origin classifier.
Infer which single 
author produced the code:
human, deepseek-ai, 
Qwen, 01-ai, bigcode, 
Gemma, Phi, meta-llama,
ibm-granite, mistralai,
OpenAI, Gemini.

Think silently to compare candidate 
profiles; 
do not reveal reasoning.
Output exactly one word: 
one of the labels above.

Use discriminative,
code-level signals:

GENERAL
- Language & libs: 
default choices (Python/JS/Java), 
PyTorch vs. TF/JAX, 
numpy/pandas usage,
pytest vs. unittest.
- Structure:
module/import ordering,
helper naming density,
dataclass/typing/async usage,
error-handling patterns.
- Textual:
identifier casing/length, 
docstring style 
(Google/NumPy/ReST/none),
comment tone,
message phrasing.
- Artifacts: template scaffolds, 
README-like headers, 
tutorial narrations, 
synthetic examples, 
placeholder vars.

CANDIDATE PROFILES 
(heuristics; match by best fit):
- OpenAI: PEP8-clean Python,
f-strings, 
type hints moderate, 
Google/NumPy docstrings, 
careful edge-case checks.
- Gemini: tendency to verbose
instructional comments,
TF/JAX-friendly snippets,
longer docstrings with bullets.
- meta-llama: PyTorch-first 
demos, 
concise comments, 
torchvision/transformers idioms, 
manual seed-setting.
- mistralai: compact Python, 
minimal ceremony, 
itertools/collections use,
terse error messages/tests.
- Qwen: occasional 
bilingual tokens/comments,
pandas/NumPy-heavy utilities,
explicit dtype handling.
- deepseek-ai: 
performance-leaning tweaks,
vectorization/numba hints,
assert-style sanity checks.
- 01-ai: straightforward 
baseline patterns, 
minimal comments, 
direct loops over abstractions.
- bigcode: repository/tooling 
scaffolds, 
license headers or 
codegen-ready templates,
typed stubs.
- Gemma: JAX/Flax hints or TF ops,
functional style utilities, 
explicit shapes in comments.
- Phi: didactic 
step-by-step snippets,
simple class wrappers, 
explicit prints/logs for tracing.
- ibm-granite:
enterprise-style structure, 
logger configuration blocks,
clear exceptions/messages.
- human: mixed 
styles/inconsistencies, 
partial impls/TODOs, 
pragmatic hacks,
commented-out experiments.

If uncertain, 
choose the single most 
plausible author by 
strongest profile match.

Code:
```{code}```


\end{lstlisting}
\end{tcolorbox}

\begin{tcolorbox}[
  colback=green!15,
  colframe=green!20,
  title=Task 3 CoT prompt,
  fonttitle=\bfseries,
  coltitle=black,
  boxrule=0.3pt,
  arc=1mm,
  enhanced,
  breakable
]
\begin{lstlisting}[breaklines=true]
You are a code-origin classifier.
Infer whether a given code snippet was written by a Human, an AI assistant, Hybridly generated, or generated by an Adversarial model trained to mimic humans and fool detectors.
Reason silently and do NOT reveal your reasoning.
Output exactly one word: Human, AI, Hybrid, or Adversarial.

When deciding, consider (but do not list) signals such as:
Naming: verbosity/descriptiveness, camelCase/snake_case consistency, oddly generic names.
Comments/docstrings: density, uniform template-like phrasing, section headers, per-function summaries.
Type hints and annotations: pervasive, perfectly consistent typing vs. ad-hoc/mixed usage.
Structure/style: consistent formatter fingerprints (Black/Prettier-like), tidy import grouping, exhaustive edge-cases, spotless spacing;
versus idiosyncratic style, small inconsistencies, TODO/WIP markers.
Boilerplate/templates: license blocks, README-like headers, instructional comments, scaffolded regions, placeholder variables (foo/bar), repeated patterns that look auto-generated.
API/library usage: unused imports, overly defensive patterns, verbose step-by-step code where a library call would suffice, redundant checks, generic logging.
Testing/examples: synthetic data, contrived examples, consistent pedantic error messages.
Artifacts of AI text: phrases like "This function...", "The following code...", or tutorial-like narration, or code that mirrors common LLM examples.
Human artifacts: partial implementations, inline hacks, commented-out experiments, domain-specific shortcuts, mismatched styles from multiple edits, quick-and-dirty error handling.

If uncertain, choose the more likely origin using the balance of signals.
Again: respond with a single word only.

Classify the origin of this code as Human, AI, Hybrid, or Adversarial.
Return exactly one word: Human, AI, Hybrid, or Adversarial.

Code:
```{code}```
\end{lstlisting}
\end{tcolorbox}

\section{Task 1 Detailed Performance}
\label{appx:task_1_performance}

\begin{figure*}
    \centering
    \includegraphics[width=\linewidth]{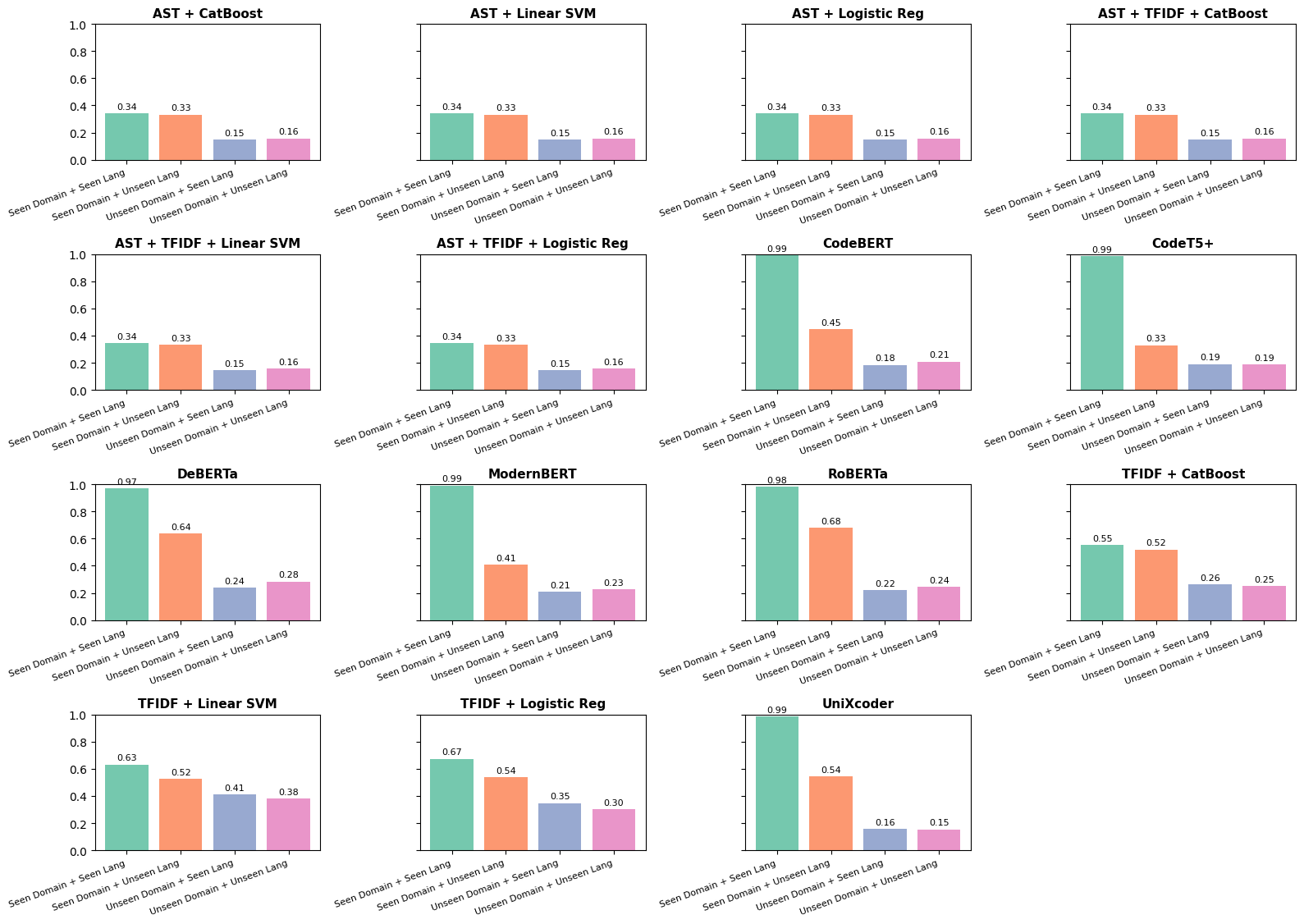}
    \caption{\textbf{Task 1 (Robust Binary Classification):}  performance evaluation  of the detectors.}
    \label{fig:task_1_modelwise}
\end{figure*}

\Cref{fig:task_1_modelwise} illustrates that the domain shift has a greater impact than the language shift. Models perform better when the programming language changes, as long as the code remains within the seen domain of algorithmic problems. Notably, performance degrades markedly on unseen domains. In fact, when both the domain and the language are unseen, performance is no worse than when only the domain is unseen, suggesting that the shift in the domain outweighs the task's overall difficulty.

Another key observation is that deep learning models exhibit strong overfitting to the training settings: they achieve near-perfect accuracy in identifying machine-generated code when both the domain and programming language match those in the training data. However, their performance drops sharply when faced with out-of-distribution examples. It persists even when the domain remains seen, but the programming language is unseen, which is a much easier setting. In contrast, the performance gap for classical models is not that large in these settings.


\begin{figure}
\centering
\includegraphics[width=0.85\linewidth]{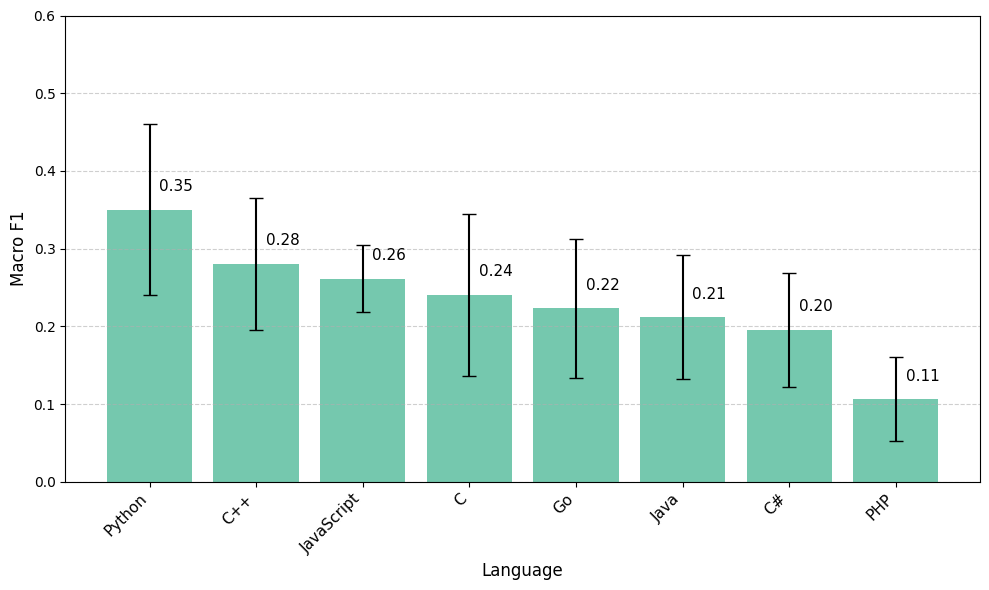}
\caption{\textbf{Task 1 (Robust Binary Classification):}  detectors' language-wise performance.}
\label{fig:task_1_languages}
\end{figure}

When comparing the programming language-wise performance, shown in \Cref{fig:task_1_languages}, it is clear that the models generally perform better when the input is Python, which is expected since Python is the language most prevalent in the training set.
For other programming languages, performance is quite similar, with the exception of PHP, which syntactically differs from the rest and, thus yielding the lowest performance.

\begin{figure*}
    \centering
    \includegraphics[width=0.85\linewidth]{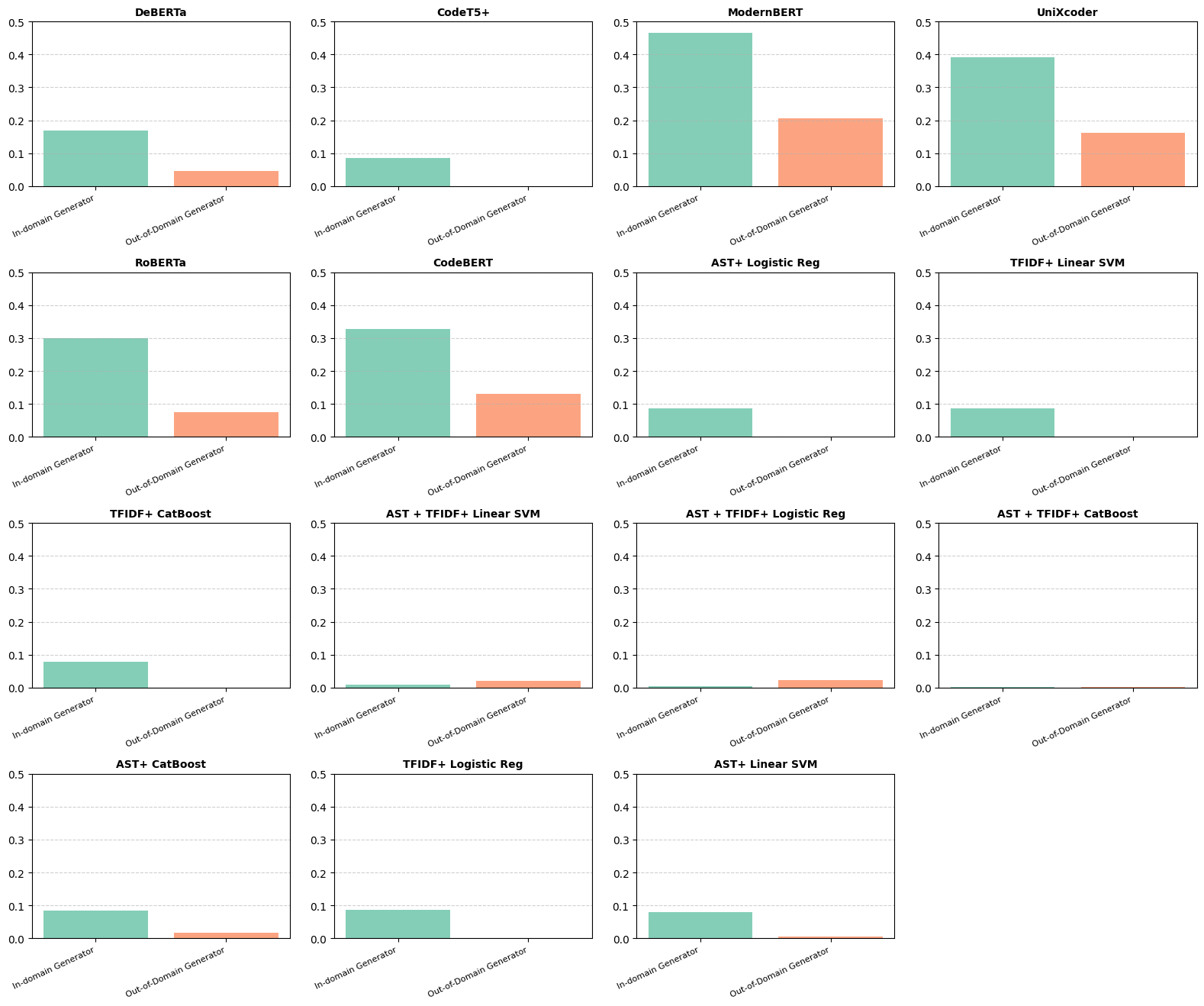}
    \caption{\textbf{Task 2 (Model Family Attribution):} performance evaluation of the detectors.}
    \label{fig:task_2_modelwise}
\end{figure*}

\subsection{SVM Performance Explained}
\label{appx:svm}
To understand why the TF-IDF SVM outperformed other baselines, we examined which n-grams most strongly indicated human vs. AI code. For interpretability, we back-projected the SVM coefficients from the SVD latent space into the original TF-IDF feature space and ranked tokens by their contributions.

We found clear stylistic differences. AI-generated code often uses verbose, prompt-echoing identifiers like \emph{answer}, \emph{output}, \emph{result}, \emph{tests}, and \emph{index}. In contrast, human-written code tended to use shorter, organic identifiers like \emph{li}, \emph{nums}, \emph{pos}, \emph{a1}, and \emph{cur}. These patterns are not tied to any specific programming language, which helps explain why the SVM generalizes well to unseen languages: the classifier captures stylistic regularities in human vs.\ AI code rather than language-specific syntax.


\section{Task 2 Detailed Performance}
\label{appx:task_2_performance}
\Cref{fig:task_2_modelwise} shows that classical machine learning models achieve nearly zero macro F1-score on out-of-domain generators, indicating a severe lack of generalization. Among these models, TF-IDF consistently yields the best performance, outperforming AST-based representations and the combination of TF-IDF and AST features. By contrast, deep learning models exhibit significantly better generalization capabilities. Among them, ModernBERT achieves the highest performance across in-domain and out-of-domain generators.

\begin{figure*}
    \centering
    \includegraphics[width=\linewidth]{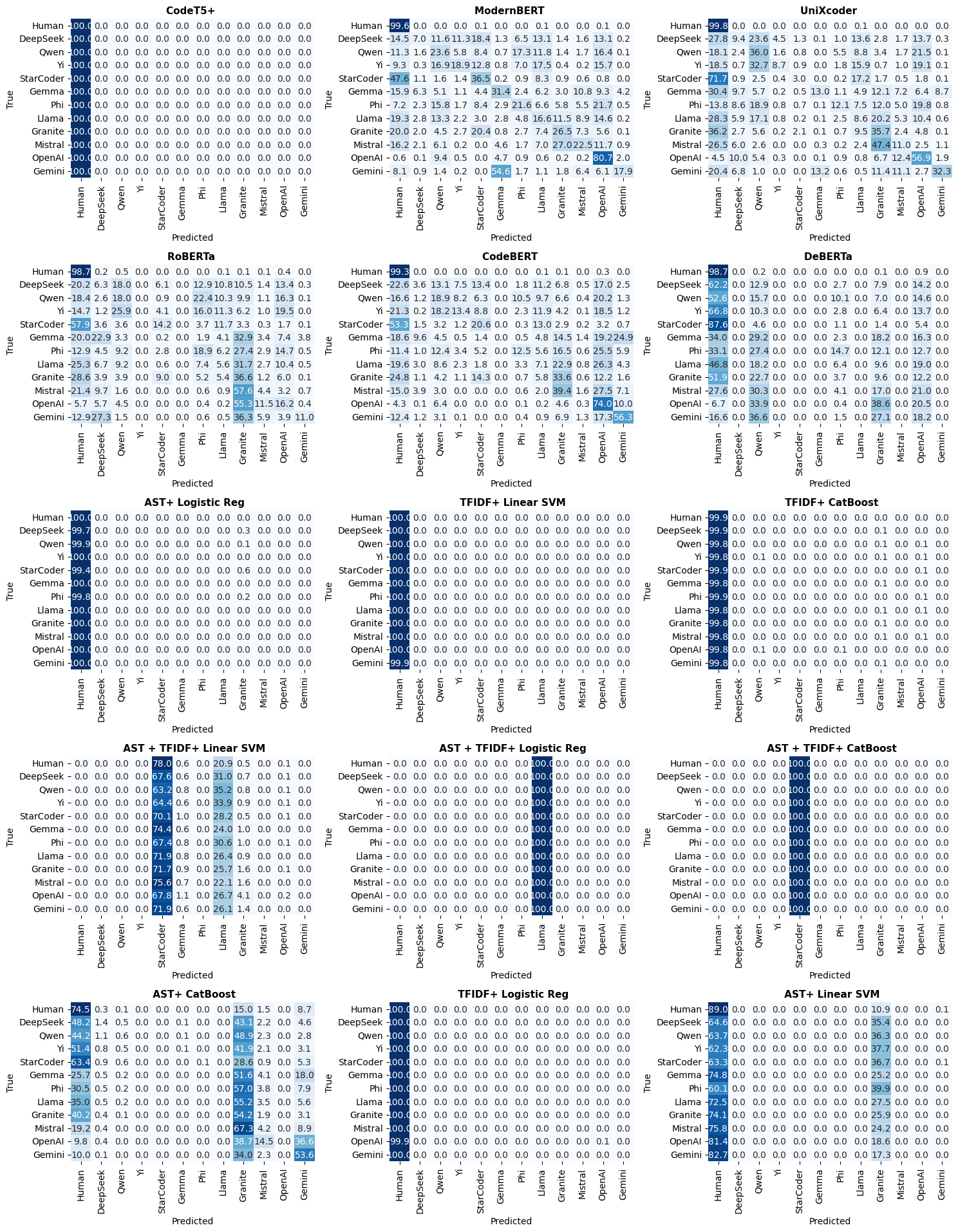}
    \caption{\textbf{Task 2 (Model Family Attribution):} confusion matrices for detectors.}
    \label{fig:task_2_confusion}
\end{figure*}

The low macro F1-score of classical models on Task 2 is further illustrated in \Cref{fig:task_2_confusion}: these models tend to predict a single dominant class, resulting in poor performance across minority classes. Similarly, among deep learning models, CodeT5+ also shows a strong bias towards predicting a single class. In contrast, the other deep learning models, particularly ModernBERT are less biased.




\section{Task 3 Detailed Performance}
\label{appx:task_3_performance}
\begin{figure*}
    \centering
    \includegraphics[width=0.8\linewidth]{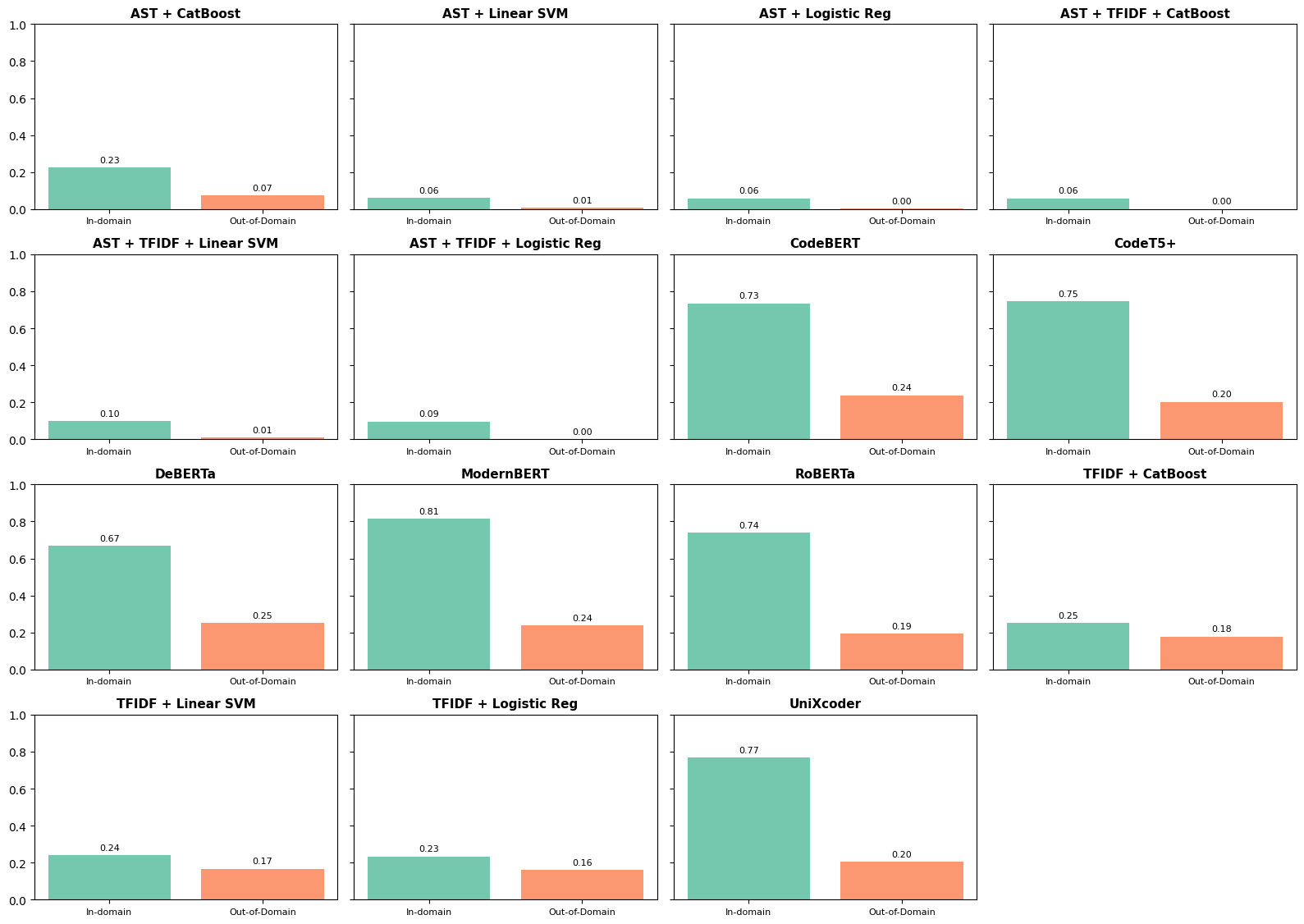}
    \caption{\textbf{Task 3 (Fine-Grained Human-Machine
Classification)}: performance evaluation of the detectors.}
    \label{fig:task_3_modelwise}
\end{figure*}

From \Cref{fig:task_3_modelwise}, we observe that only the CatBoost model can utilize AST features to work at least in-domain settings. 
Other classical models fail shortly when trying to use these features. In case of TF-IDF, on the other hand, all classical models have demonstrated comparable performance for both in-domain and out-of-domain data. Deep learning models, similarly to Task 1, achieve good in-domain performance (over 71\%) while performing poorly out-of-domain (below 25\%).

\begin{figure*}
    \centering
    \includegraphics[width=\linewidth]{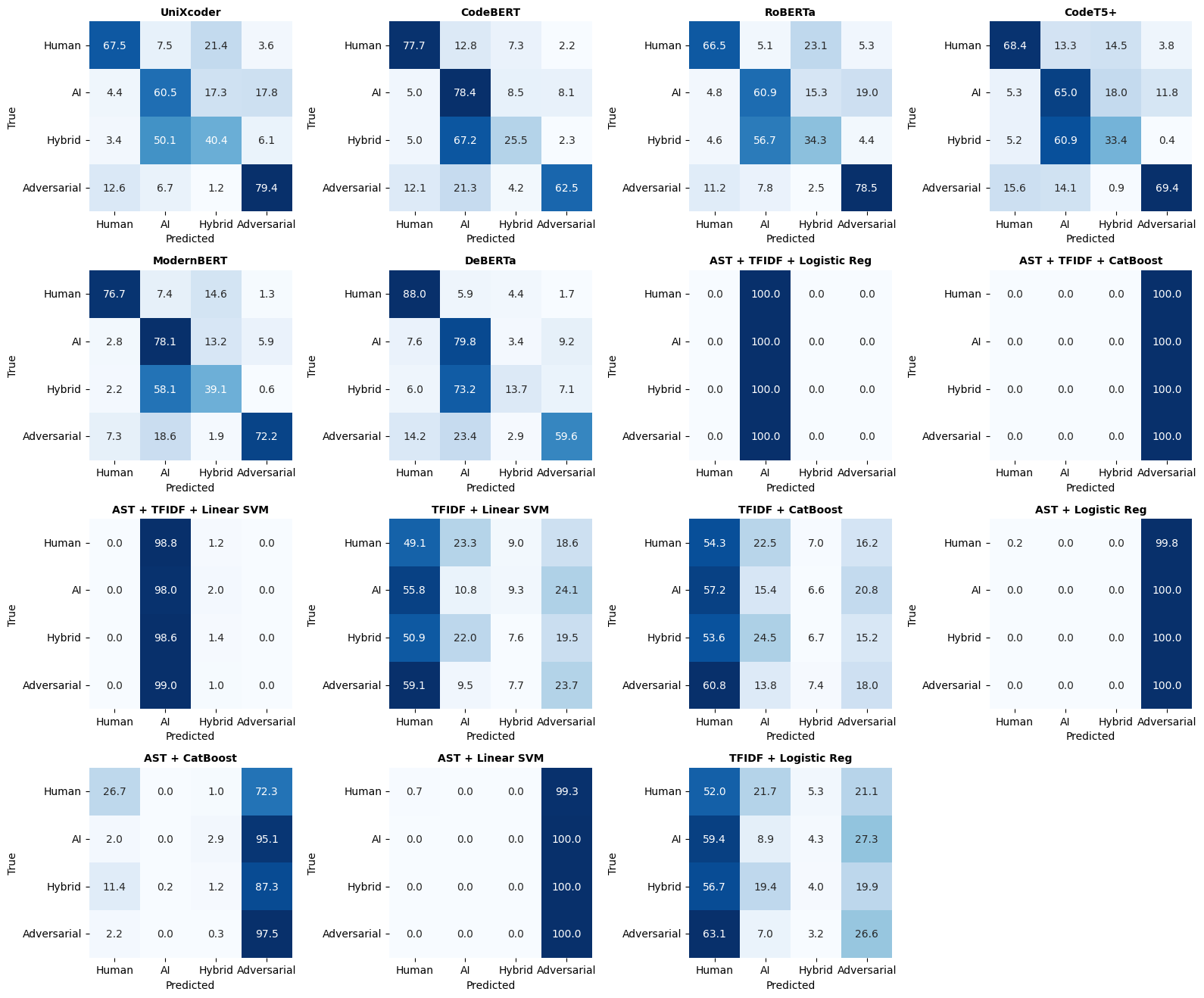}
    \caption{\textbf{Task 3 (Fine-Grained Human-Machine
Classification)}: confusion matrices for detectors.}
    \label{fig:task_3_modelwise_confusion}
\end{figure*}

When analyzing the confusion matrices for the models (\Cref{fig:task_3_modelwise_confusion}), it is obvious that the classical models barely learned anything: they were mainly predicting a single class. Deep learning models, in contrast, tend to learn correct class assignments. Interestingly, adversarial code samples are often misclassified as human- or AI-written, highlighting that the adversarial generation achieved its goals.
Additionally, hybrid cases are being misclassified with AI-generated ones, as hybrid generation also involves code rewriting that can drastically alter the initial human-written code structure.

\section{Examples of Model Failures}
\label{appx:errors}
\Cref{tab:explanations_1,tab:explanations_2,tab:explanations_3} showcase where all baselines fail. In Task 1, models often fail on very short code snippets. Boilerplate code is also frequently mislabeled as AI-generated, likely because such patterns are common in Code-LM training data, leading models to misclassify them as AI output. For Task 2, we omit predicted classes since models rarely agree; each typically assigns a different class to the same input, revealing task difficulty.
In Task 3, errors again focus on boilerplate code. Also, misclassifications mostly occur among similar categories, AI-generated, hybrid, and adversarial, indicating that models struggle to discern fine-grained distinctions between them.

\lstdefinestyle{codeStyle}{
  basicstyle=\ttfamily\small,
  frame=single,
  breaklines=true,
  columns=fullflexible,
  keepspaces=true,
  showstringspaces=false,
   keywordstyle=\color{blue},
  commentstyle=\color{gray!60},
  stringstyle=\color{teal},
}
\begin{table*}[!t]
\centering
\begin{adjustbox}{width=\textwidth}
\begin{tabular}{|p{18cm}|c|c|p{5cm}|}
\toprule
\textbf{Code} & \textbf{True Label} & \textbf{Predicted Label} & \textbf{Possible explanation}  \\
\midrule
\begin{lstlisting}[language=Python,style=codeStyle]
def max(a, b):
    return a if a > b else b


def min(a, b):
    return a if a < b else b


class StringHash:
    def __init__(self, lst):
        """two mod to avoid hash crush"""
        # use two class to compute is faster!!!
        self.n = len(lst)
        self.p = random.randint(26, 100)
        self.mod = random.randint(10 ** 9 + 7, 2 ** 31 - 1)
        self.pre = [0] * (self.n + 1)
        self.pp = [1] * (self.n + 1)
        for j, w in enumerate(lst):
            self.pre[j + 1] = (self.pre[j] * self.p + w) % self.mod
            self.pp[j + 1] = (self.pp[j] * self.p) % self.mod
        return

    def query(self, x, y):
        """range hash value index start from 0"""
        # assert 0 <= x <= y <= self.n - 1
        if y < x:
            return 0
        # with length y - x + 1 important!!!
        ans = (self.pre[y + 1] - self.pre[x] * self.pp[y - x + 1]) % self.mod
        return ans


class Solution:
    def countPrefixSuffixPairs(self, words: List[str]) -> int:
        ans = 0
        st = "".join(words)
        sh1 = StringHash([ord(w) - ord("a") for w in st])
        sh2 = StringHash([ord(w) - ord("a") for w in st])
        pre = defaultdict(int)
        length = 0
        for word in words:
            m = len(word)
            for i in range(1, m+1):
                prefix = (sh1.query(length, length+i-1), sh2.query(length, length+i-1), i)
                suffix = (sh1.query(length+m-1-i+1, length + m-1), sh2.query(length+m-1-i+1, length + m-1), i)
                if prefix == suffix:
                    ans += pre[prefix]

            prefix = (sh1.query(length, length + m - 1), sh2.query(length, length + m - 1), m)
            pre[prefix] += 1
            length += m
        return ans
\end{lstlisting}
& Human-Written & AI-Generated & 
Dosctrings in StringHash do not look human-written. That may be the case that StringHash is just a Boilerplate implementation
\\
\midrule
\begin{lstlisting}[language=Python,style=codeStyle]
s = input().strip()

x = int(input().strip())

if (s=="ABC" and x<2000) or (s=="ARC" and x<2800) or (s=="AGC" and x>=1200):

    print('yes')

else:

    print('no')
\end{lstlisting}
& Human-Written& AI-Generated  &
Short code with not enough signal for classification
\\
\midrule
\begin{lstlisting}[language=Python,style=codeStyle]
N,*A=map(int,open(0).read().split())

T=[*zip(A,A[4:]+A[3:4])]

print(sum(max(0,a+b-N)for a,b in T),sum(map(min,T)))
\end{lstlisting}
& Human-Written& AI-Generated  &
Short code with not enough signal for classification
\\
\bottomrule
\end{tabular}
\end{adjustbox}
\caption{\textbf{Task 1 (Robust Binary Classification):} examples of model misclassification and their possible explanations.}
\label{tab:explanations_1}
\end{table*}

\begin{table*}[!t]
\centering
\begin{adjustbox}{width=\textwidth}
\begin{tabular}{|p{20cm}|c|}
\toprule
\textbf{Code} & \textbf{Model Family} \\
\midrule
\begin{lstlisting}[language=Java,style=codeStyle]
import javax.crypto.Cipher;
import javax.crypto.KeyGenerator;
import javax.crypto.SecretKey;
import javax.crypto.spec.IvParameterSpec;
import javax.crypto.spec.SecretKeySpec;
import java.security.KeyStore;
import java.security.spec.PKCS8EncodedKeySpec;
import javax.security.auth.callback.*;
import javax.security.auth.Subject;
import javax.security.auth.spi.*;
import java.util.*;
import java.nio.file.*;
import java.security.*;
import java.util.logging.*;

public class IoTSecurityFramework {

    private static final String ALGORITHM = "AES/CBC/PKCS5Padding";
    private static final Logger logger = Logger.getLogger(IoTSecurityFramework.class.getName());
    private SecretKey secretKey;
    private IvParameterSpec ivSpec;

    public IoTSecurityFramework() throws Exception {
        // Generate symmetric key for AES
        KeyGenerator keyGen = KeyGenerator.getInstance("AES");
        keyGen.init(128); // Key size
        secretKey = keyGen.generateKey();
        ivSpec = new IvParameterSpec(new byte[16]); // Initialization vector
    }

    public byte[] encrypt(String data) throws Exception {
        Cipher cipher = Cipher.getInstance(ALGORITHM);
        cipher.init(Cipher.ENCRYPT_MODE, secretKey, ivSpec);
        return cipher.doFinal(data.getBytes());
    }

    public String decrypt(byte[] encryptedData) throws Exception {
        Cipher cipher = Cipher.getInstance(ALGORITHM);
        cipher.init(Cipher.DECRYPT_MODE, secretKey, ivSpec);
        return new String(cipher.doFinal(encryptedData));
    }

    public boolean authenticateDevice(String deviceId, String sessionToken) {
        // Mock authentication
        return "validDeviceId".equals(deviceId) && "validSessionToken".equals(sessionToken);
    }
    public void roleBasedAccessControl(String role) throws SecurityException {
        // Simple RBAC check
        if (!"admin".equals(role)) {
            throw new SecurityException("Access Denied for role: " + role);
        }
    }
    public void logAccessEvent(String message) {
        logger.info(message);
    }
    public void monitorForIntrusion() {
        // Placeholder for intrusion detection mechanism
        System.out.println("Monitoring for unauthorized access...");
    }
    public boolean secureCodingGuide() {
        // Placeholder for guide return
        System.out.println("Guidelines: Validate inputs, Use prepared statements, Error handling.");
        return true;
    }
    public static void main(String[] args) throws Exception {
        IoTSecurityFramework securityFramework = new IoTSecurityFramework();
        // Sample test
        String originalData = "Sensitive IoT Data";
        byte[] encryptedData = securityFramework.encrypt(originalData);
        String decryptedData = securityFramework.decrypt(encryptedData);
        securityFramework.logAccessEvent("Data encrypted and decrypted successfully.");
        System.out.println("Decrypted data: " + decryptedData);
        securityFramework.monitorForIntrusion();
        try {
            securityFramework.roleBasedAccessControl("user");
        } catch (SecurityException e) {
            securityFramework.logAccessEvent(e.getMessage());
        }

        securityFramework.secureCodingGuide();
    }
}
\end{lstlisting}
& Granite
\\
\midrule
\begin{lstlisting}[language=Python,style=codeStyle]
def data_gen_args(self, context: str) -> dict:
        if context == 'train':
            return dict(
                horizontal_flip=self.horizontal_flip,
                vertical_flip=self.vertical_flip,
                image_size=self.crop_size
            )
        return dict(image_size=self.crop_size)
\end{lstlisting}
& Mistral\\
\bottomrule
\end{tabular}
\end{adjustbox}
\caption{\textbf{Task 2 (Model Family Attribution):} code snippets that caused errors.}
\label{tab:explanations_2}
\end{table*}

\begin{table*}[!t]
\centering
\begin{adjustbox}{width=\textwidth}
\begin{tabular}{|p{18cm}|c|c|p{5cm}|}
\toprule
\textbf{Code} & \textbf{True Label} & \textbf{Predicted Label} & \textbf{Possible explanation}  \\
\midrule
\begin{lstlisting}[language=Python,style=codeStyle]
import pygame
import sys
import time

# Constants
SCREEN_WIDTH = 1026
SCREEN_HEIGHT = 700
WINNING_SOUND_FILE = 'Sound Effects/smb_stage_clear.wav'
BACKGROUND_IMAGE_FILE = "Images/Background2.jpg"
FONT_FILE = 'font.ttf'
WIN_TEXT = 'YOU WON!'
WIN_TEXT_COLOR = (50, 205, 50)
FONT_SIZE = 26
GAME_DURATION = 6  # seconds

class Game:
    """Represents the game."""

    def __init__(self):
        """Initializes the game."""
        pygame.init()
        self.screen = pygame.display.set_mode((SCREEN_WIDTH, SCREEN_HEIGHT))
        pygame.display.set_caption("Super Python Bros.")
        self.clock = pygame.time.Clock()
        self.winning_sound = self.load_winning_sound()
        self.background_image = self.load_background_image()
        self.win_text = self.render_win_text()

    def load_winning_sound(self) -> pygame.mixer.Sound | None:
        """Loads the winning sound effect."""
        try:
            return pygame.mixer.Sound(WINNING_SOUND_FILE)
        except pygame.error as e:
            print(f"Error loading winning sound: {e}")
            return None

    def load_background_image(self) -> pygame.Surface | None:
        """Loads the background image."""
        try:
            return pygame.image.load(BACKGROUND_IMAGE_FILE).convert()
        except pygame.error as e:
            print(f"Error loading background image: {e}")
            return None

    def render_win_text(self) -> pygame.Surface:
        """Renders the win text."""
        font = pygame.font.Font(FONT_FILE, FONT_SIZE)
        return font.render(WIN_TEXT, False, WIN_TEXT_COLOR)

    def build_win_menu(self) -> None:
        """Builds the win menu."""
        # Clear the screen with the background image
        self.screen.blit(self.background_image, (0, 0))
        # Render the win text
        self.screen.blit(self.win_text, (400, 345))
        # Update the display
        pygame.display.flip()

    def run(self) -> None:
        """Runs the game."""
        if self.winning_sound:
            self.winning_sound.play()

        end_time = time.time() + GAME_DURATION
        while time.time() < end_time:
            self.build_win_menu()

            for event in pygame.event.get():
                if event.type == pygame.QUIT:
                    pygame.quit()
                    sys.exit()
                elif event.type == pygame.KEYDOWN:
                    if event.key == pygame.K_ESCAPE:
                        pygame.quit()
                        sys.exit()

            self.clock.tick(60)

if __name__ == "__main__":
    game = Game()
    game.run()
\end{lstlisting}
& Hybrid & mainly AI-Generated, few Adversarial & 
Since it is a hybrid case, misclassification happens with two most similar cases
\\
\midrule
\begin{lstlisting}[language=Python,style=codeStyle]
package zmq.util;

// Emulates the errno mechanism present in C++, in a per-thread basis.
public final class Errno
{
    private static final ThreadLocal<Integer> local = ThreadLocal.withInitial(() -> 0);

    public int get()
    {
        return local.get();
    }

    public void set(int errno)
    {
        local.set(errno);
    }

    public boolean is(int err)
    {
        return get() == err;
    }

    @Override
    public String toString()
    {
        return "Errno[" + get() + "]";
    }
}
\end{lstlisting}
& Human-Written & All classes except Human-Written  &
Too boiler-plate like
\\
\bottomrule
\end{tabular}
\end{adjustbox}
\caption{\textbf{Task 3 (Fine-Grained Human-Machine
Classification)}: examples of model misclassification and their possible explanations.}
\label{tab:explanations_3}
\end{table*}

\section{SHAP analysis}
\label{appx:shap}
\begin{figure*}[t]
  \centering

  \begin{minipage}[t]{0.32\textwidth}
    \centering
    \includegraphics[width=\linewidth]{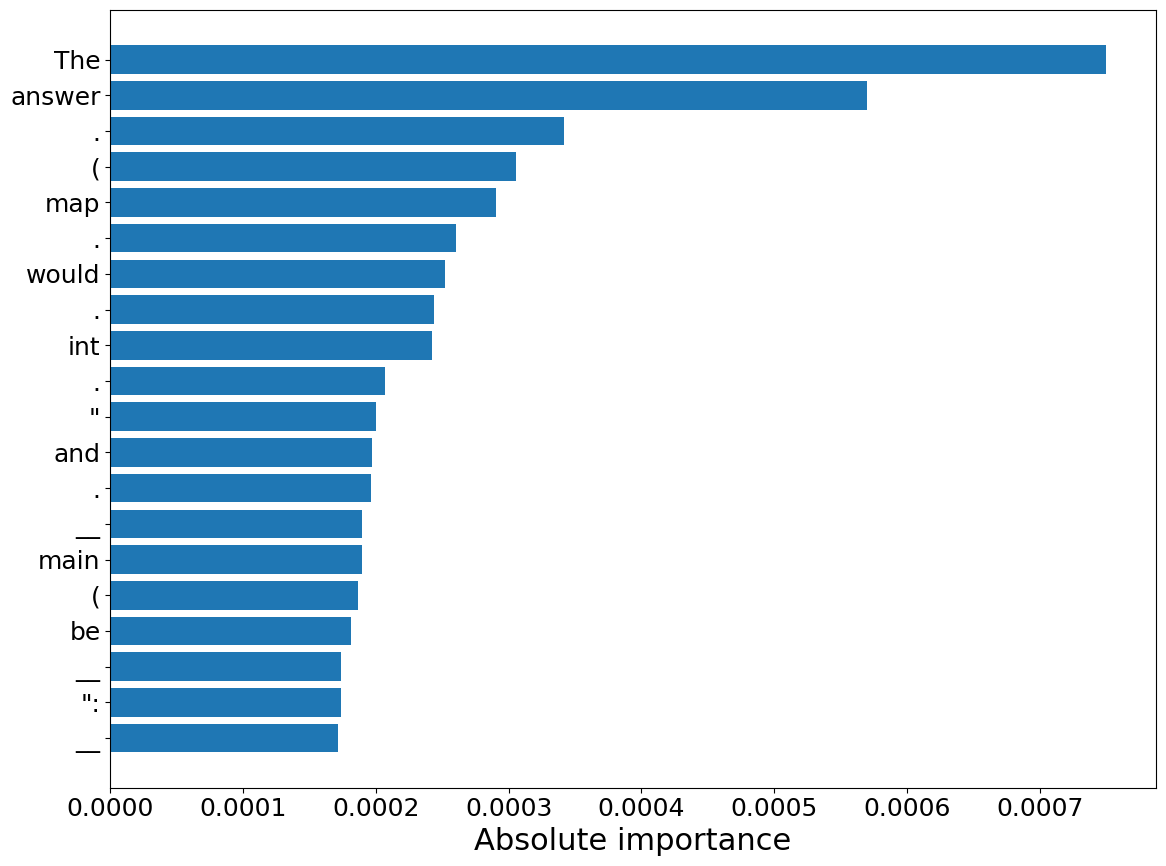}
  \end{minipage}\hfill
  \begin{minipage}[t]{0.32\textwidth}
    \centering
    \includegraphics[width=\linewidth]{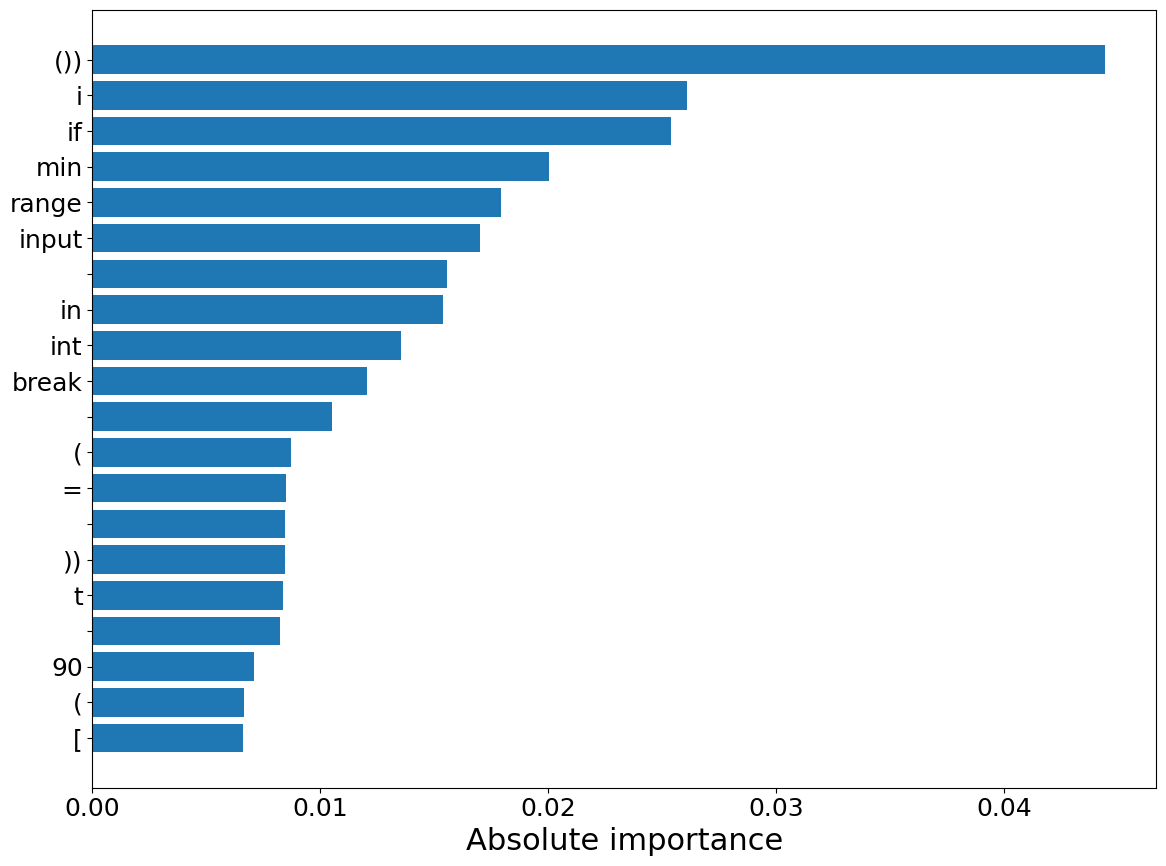}
  \end{minipage}\hfill
  \begin{minipage}[t]{0.32\textwidth}
    \centering
    \includegraphics[width=\linewidth]{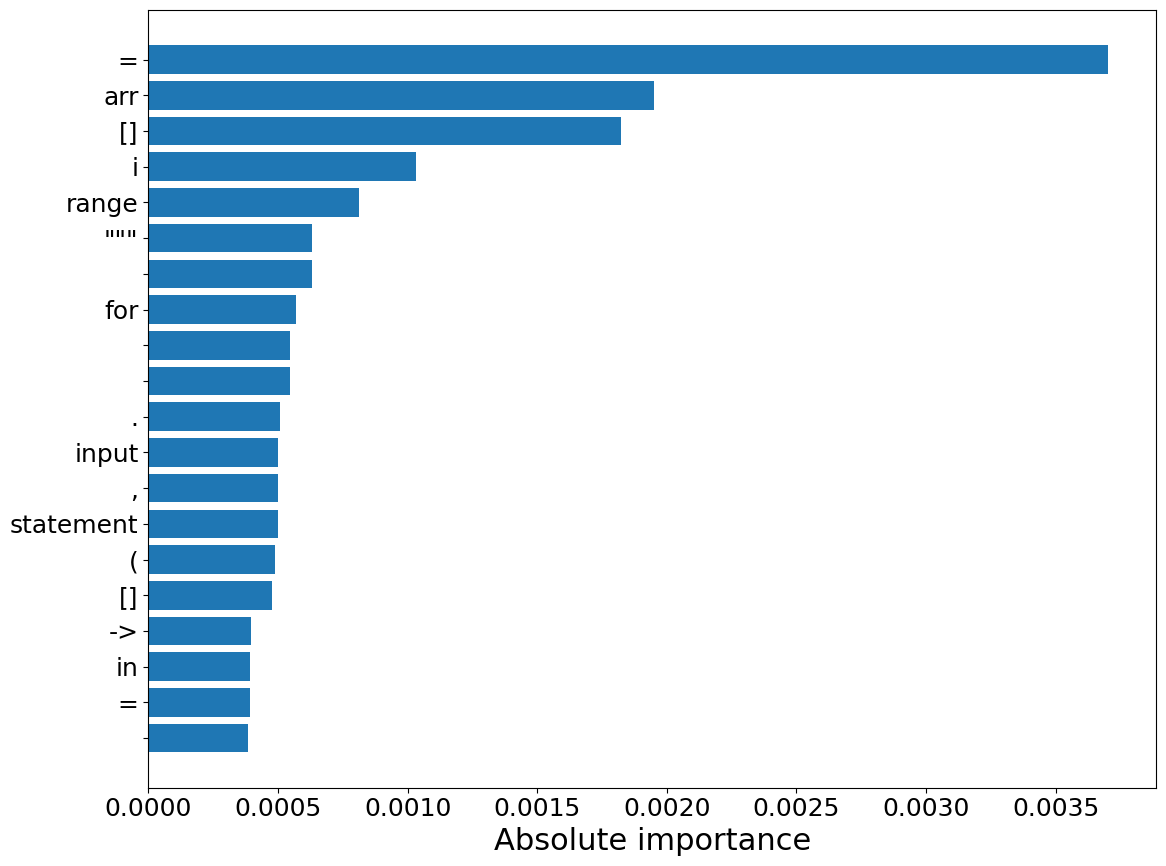}
  \end{minipage}

  {\small SHAP for correct predictions.}


  \begin{minipage}[t]{0.32\textwidth}
    \centering
    \includegraphics[width=\linewidth]{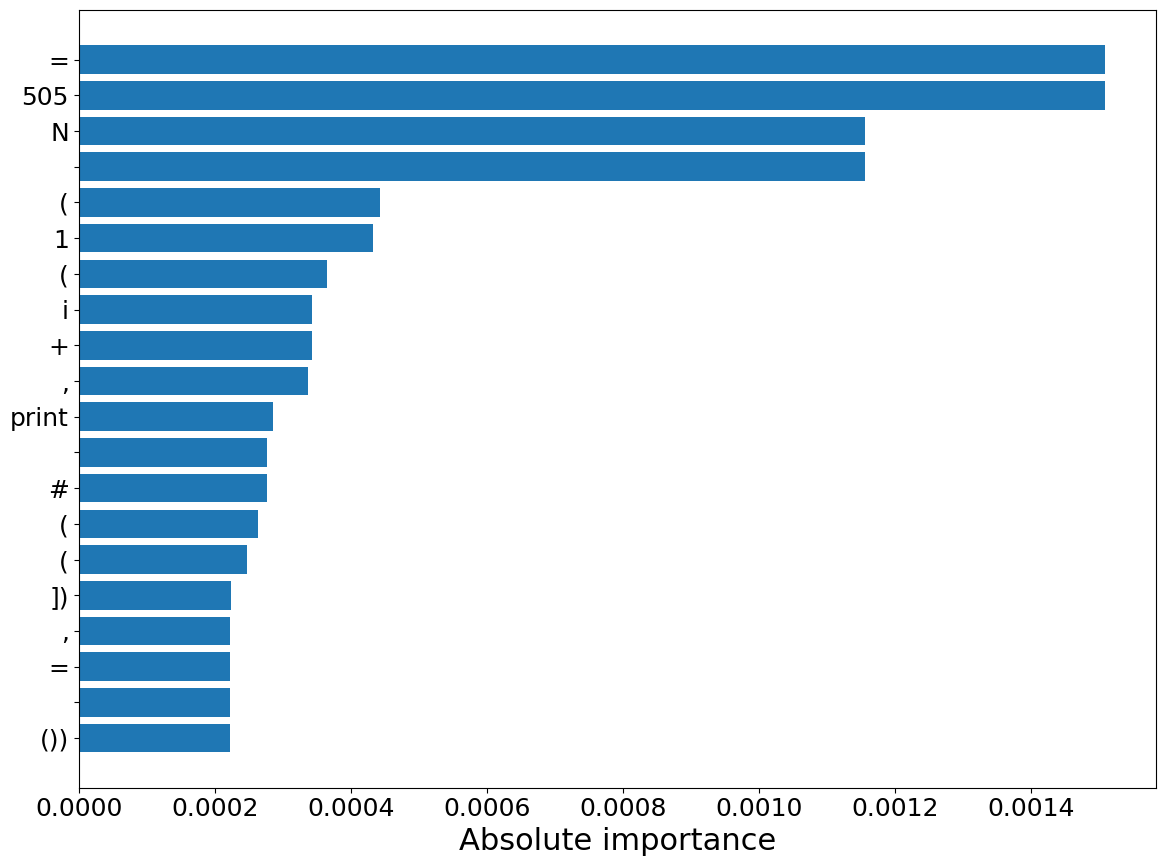}
  \end{minipage}\hfill
  \begin{minipage}[t]{0.32\textwidth}
    \centering
    \includegraphics[width=\linewidth]{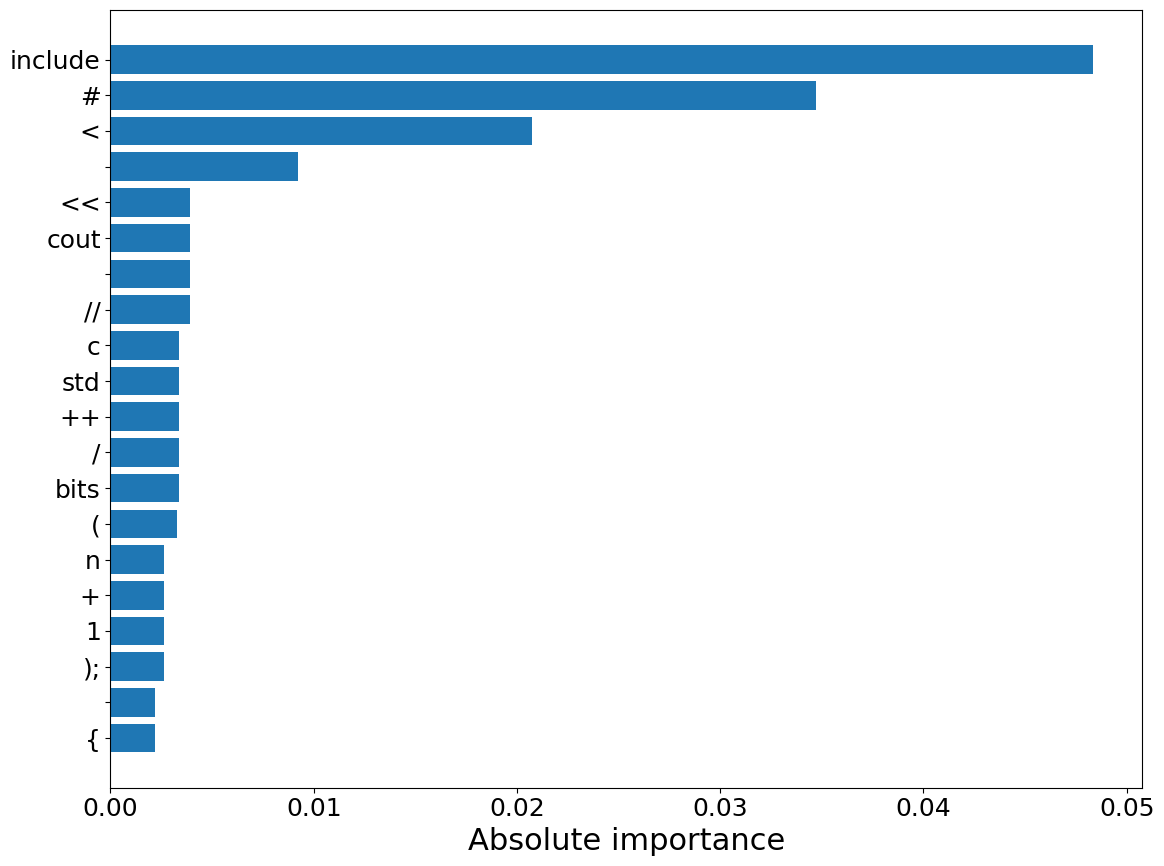}
  \end{minipage}\hfill
  \begin{minipage}[t]{0.32\textwidth}
    \centering
    \includegraphics[width=\linewidth]{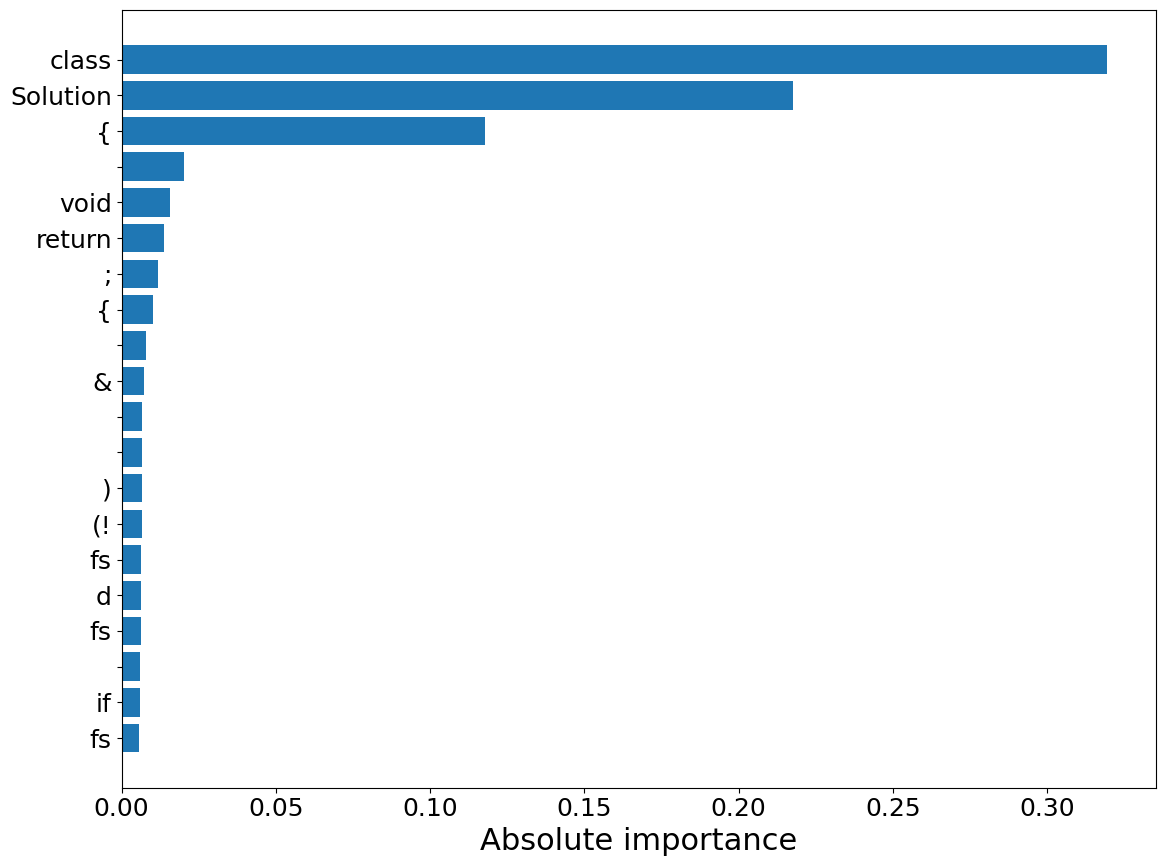}
  \end{minipage}

  {\small SHAP for incorrect predictions.}

  \caption{\textbf{Task 1 (Robust Binary Classification):} token-level SHAP visualizations comparing correct vs.\ incorrect predictions.}
  \label{fig:shap-correct-vs-incorrect}
\end{figure*}

\begin{figure*}[t]
  \centering

  \begin{minipage}[t]{0.32\textwidth}
    \centering
    \includegraphics[width=\linewidth]{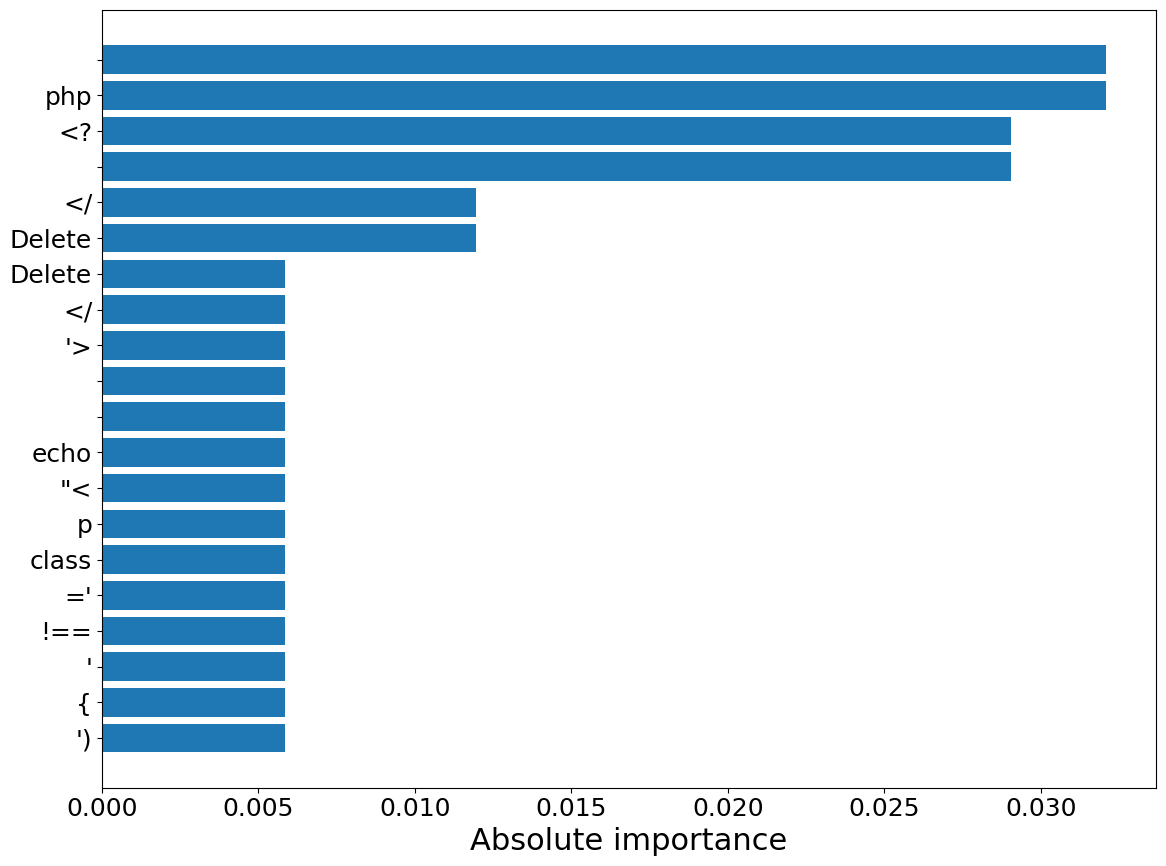}
  \end{minipage}\hfill
  \begin{minipage}[t]{0.32\textwidth}
    \centering
    \includegraphics[width=\linewidth]{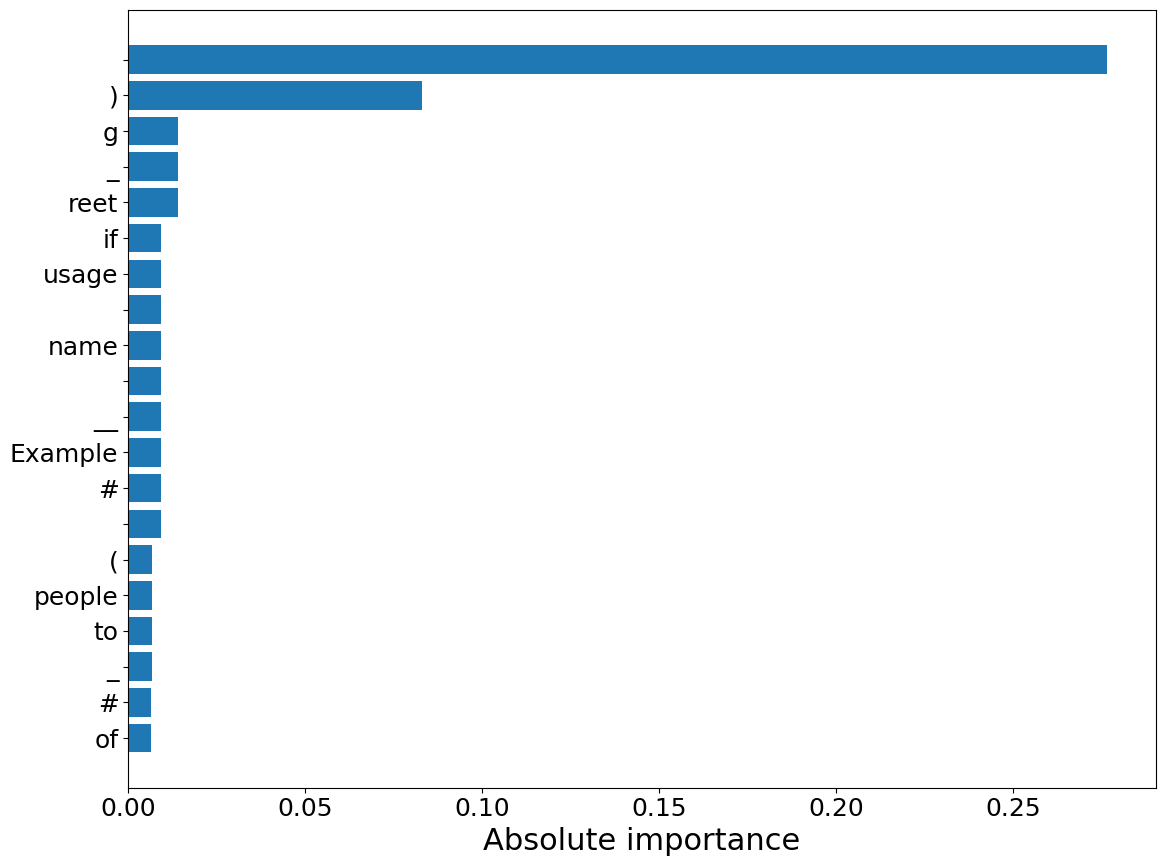}
  \end{minipage}\hfill
  \begin{minipage}[t]{0.32\textwidth}
    \centering
    \includegraphics[width=\linewidth]{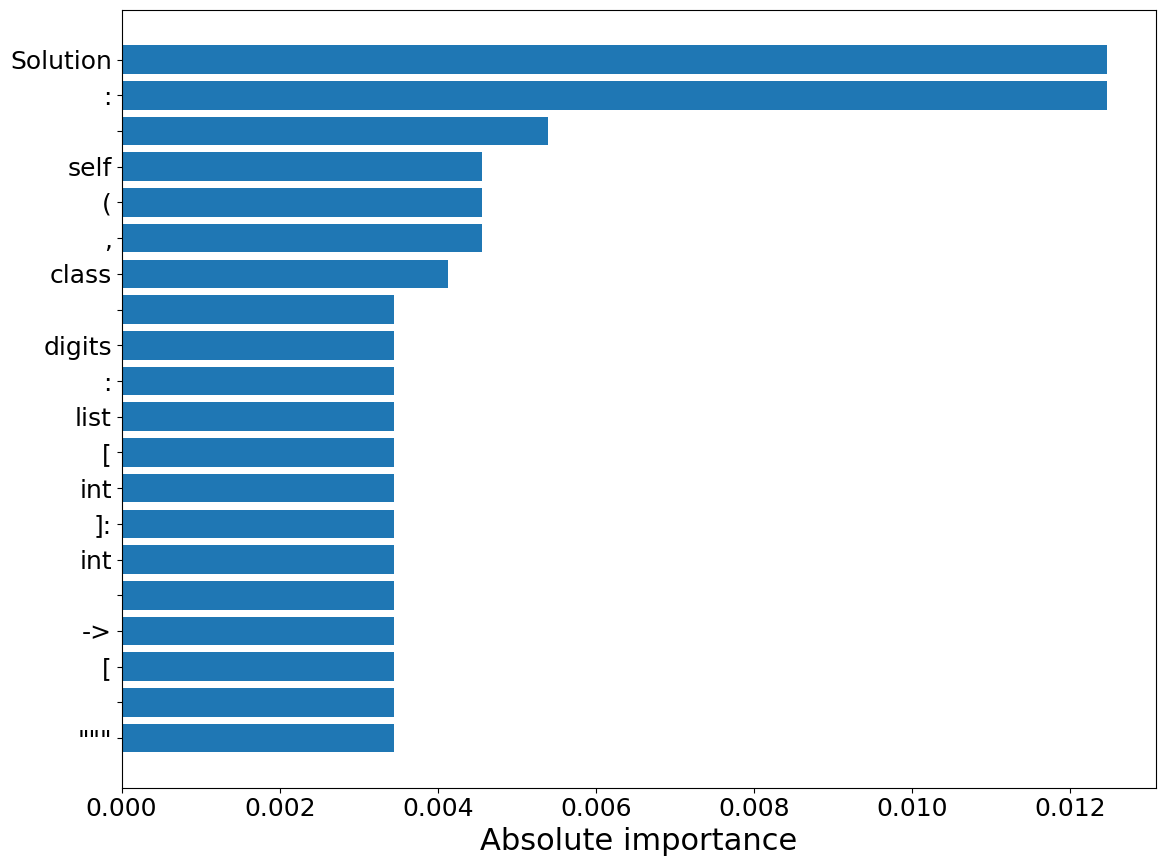}
  \end{minipage}

  {\small SHAP for correct predictions.}


  \begin{minipage}[t]{0.32\textwidth}
    \centering
    \includegraphics[width=\linewidth]{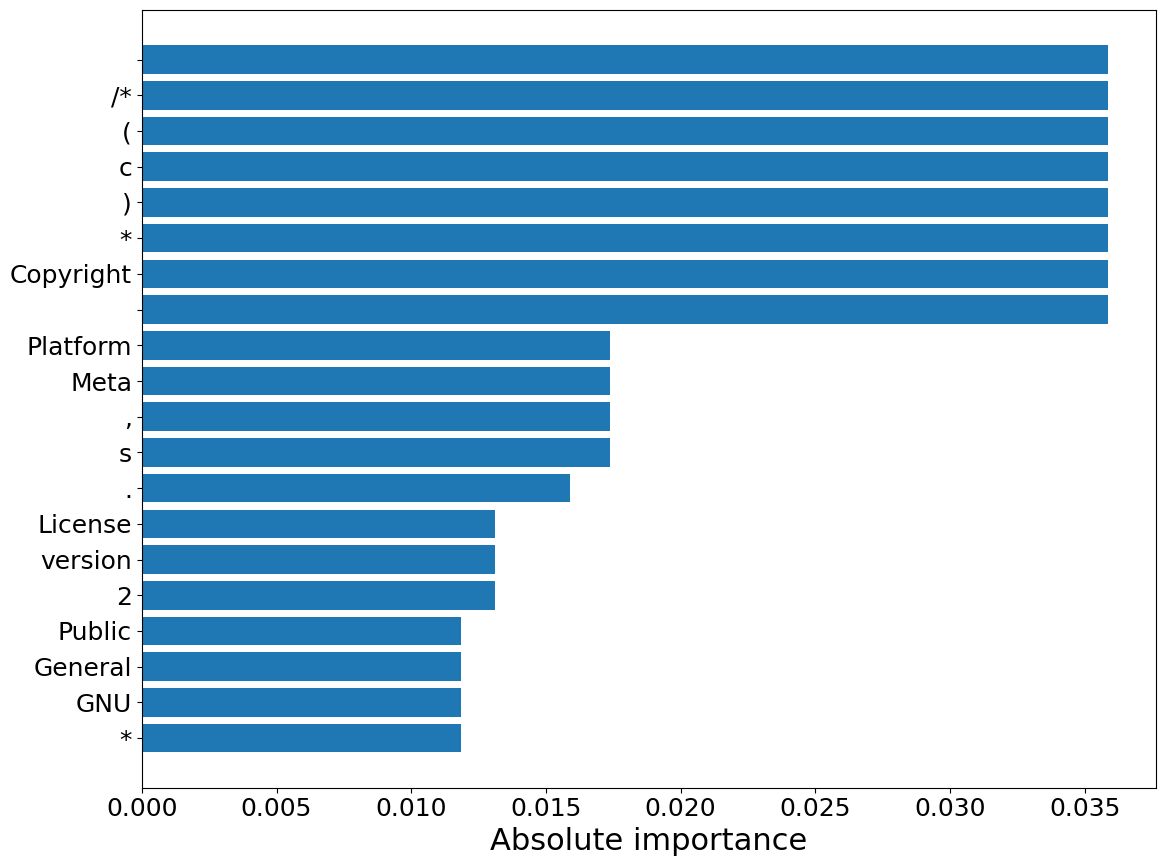}
  \end{minipage}\hfill
  \begin{minipage}[t]{0.32\textwidth}
    \centering
    \includegraphics[width=\linewidth]{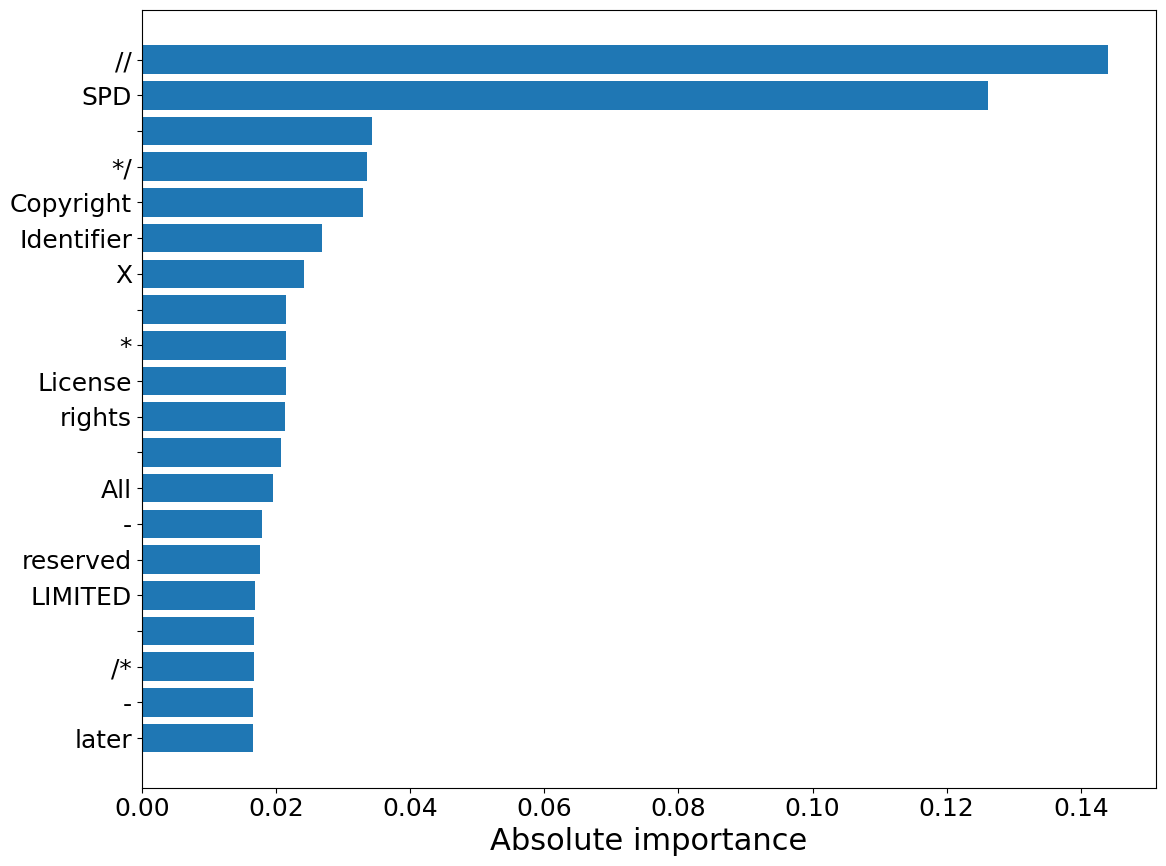}
  \end{minipage}\hfill
  \begin{minipage}[t]{0.32\textwidth}
    \centering
    \includegraphics[width=\linewidth]{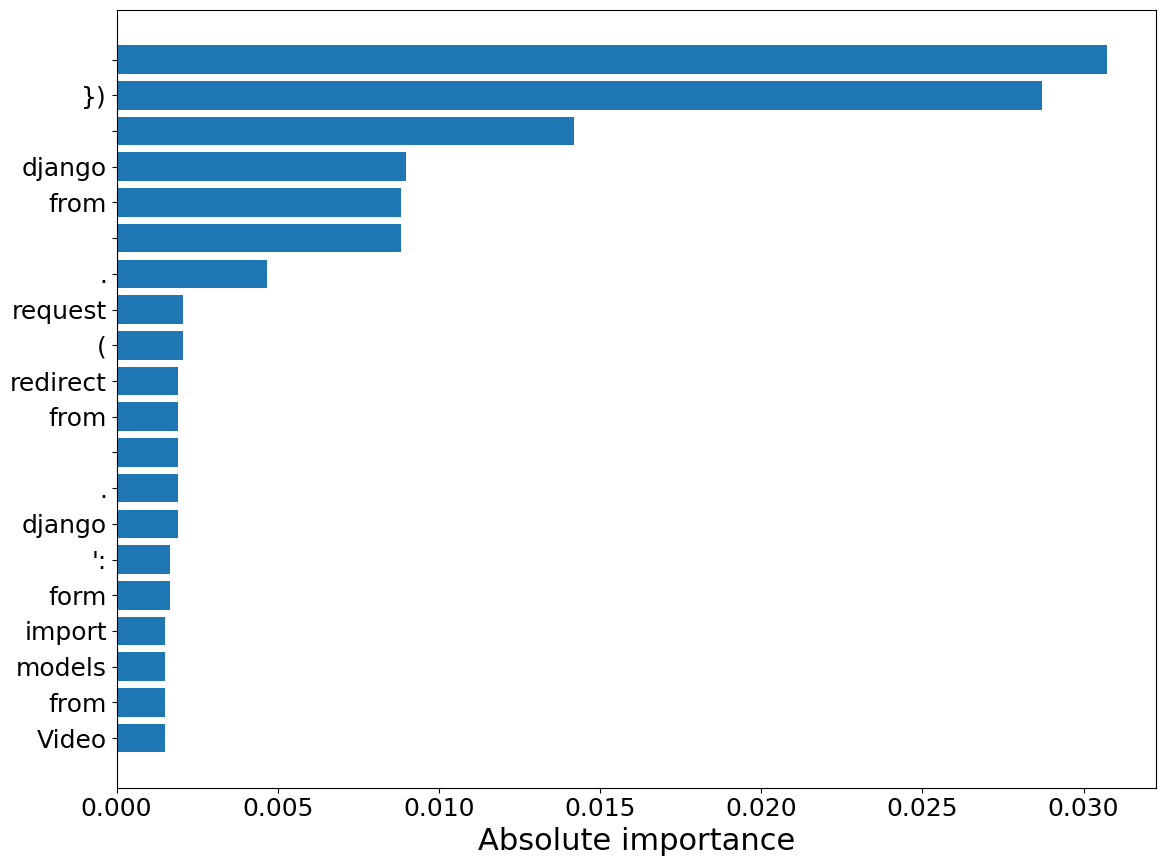}
  \end{minipage}

  {\small SHAP for incorrect predictions.}

  \caption{\textbf{Task 3 (Fine-Grained Human-Machine
Classification):} token-level SHAP visualizations comparing correct vs.\ incorrect predictions for hybrid class prediction.}
  \label{fig:shap-correct-vs-incorrect-t3}
\end{figure*}

\Cref{fig:shap-correct-vs-incorrect} compares token-level SHAP visualizations for correctly and incorrectly classified samples in Task 1, while \Cref{fig:shap-correct-vs-incorrect-t3} shows the analogous comparison for hybrid cases in Task 3. We also manually inspect additional examples; the figures show representative samples that illustrate the attribution patterns observed in a more general sense. Analysis of other tasks and classes did not yield useful information
.

\end{document}